\documentclass[10pt,journal,compsoc]{IEEEtran}
\usepackage{times}
\usepackage{soul}
\usepackage{url}
\usepackage[hidelinks]{hyperref}
\usepackage[utf8]{inputenc}
\usepackage[small]{caption}
\usepackage{graphicx}
\usepackage{amsmath}
\usepackage{amsthm}
\usepackage{amsfonts}
\usepackage{bm}
\usepackage{booktabs}
\usepackage{algorithm}
\usepackage{algorithmic}
\usepackage[switch]{lineno}
\usepackage{multirow}
\usepackage{makecell}
\usepackage{subfigure}

\usepackage[capitalize]{cleveref}
\crefname{section}{Sec.}{Secs.}
\Crefname{section}{Section}{Sections}
\Crefname{table}{Table}{Tables}
\crefname{table}{Tab.}{Tabs.}

\makeatletter
\DeclareRobustCommand\onedot{\futurelet\@let@token\@onedot}
\def\@onedot{\ifx\@let@token.\else.\null\fi\xspace}

\makeatother

\makeatletter
\def\algbackskip{\hskip-\ALG@thistlm}
\newcommand{\nosemic}{\renewcommand{\@endalgocfline}{\relax}}
\newcommand{\dosemic}{\renewcommand{\@endalgocfline}{\algocf@endline}}
\let\oldnl\nl
\newcommand{\nonl}{\renewcommand{\nl}{\let\nl\oldnl}}
\makeatother

\allowdisplaybreaks[3]

\begin{document}

\title{FedMHO: Heterogeneous One-Shot Federated Learning Towards Resource-Constrained Edge Devices}

\author{Dezhong~Yao,
        Yuexin Shi, 
        Tongtong Liu, 
        Zhiqiang Xu
    \thanks{
        \quad Dezhong Yao, Yuexin Shi, and Tongtong Liu are with the National Engineering Research Center for Big Data Technology and System, Services Computing Technology and System Lab, Cluster and Grid Computing Lab, School of Computer Science and Technology, Huazhong University of Science and Technology, Wuhan 430074, China (E-mail: \{dyao, shiyuexin, tliu\}@hust.edu.cn).}
    \thanks{
        \quad Zhiqiang Xu is with Mohamed bin Zayed University of Artificial Intelligence, UAE (E-mail: zhiqiang.xu@mbzuai.ac.ae). 
    }
    \thanks{(Corresponding author: Dezhong~Yao.)}
}

%
%

\markboth{Under review}%
{Shell \MakeLowercase{\textit{et al.}}: Bare Demo of IEEEtran.cls for Computer Society Journals}
%

\IEEEtitleabstractindextext{%
\begin{abstract}
Federated Learning (FL) is increasingly adopted in edge computing scenarios, where a large number of heterogeneous clients operate under constrained or sufficient resources. The iterative training process in conventional FL introduces significant computation and communication overhead, which is unfriendly for resource-constrained edge devices. One-shot FL has emerged as a promising approach to mitigate communication overhead, and model-heterogeneous FL solves the problem of diverse computing resources across clients. However, existing methods face challenges in effectively managing model-heterogeneous one-shot FL, often leading to unsatisfactory global model performance or reliance on auxiliary datasets. To address these challenges, we propose a novel FL framework named FedMHO, which leverages deep classification models on resource-sufficient clients and lightweight generative models on resource-constrained devices. On the server side, FedMHO involves a two-stage process that includes data generation and knowledge fusion. Furthermore, we introduce FedMHO-MD and FedMHO-SD to mitigate the knowledge-forgetting problem during the knowledge fusion stage, and an unsupervised data optimization solution to improve the quality of synthetic samples. Comprehensive experiments demonstrate the effectiveness of our methods, as they outperform state-of-the-art baselines in various experimental setups. Our code is available at~\url{https://github.com/YXShi2000/FedMHO}.
\end{abstract}

}

\maketitle

\IEEEdisplaynontitleabstractindextext

%
\IEEEpeerreviewmaketitle

\section{Introduction}
\label{section_introduction}
With the advancement and adoption of Edge Computing (EC)~\cite{wu2023topology}, edge devices continuously generate vast amounts of data~\cite{chen2022decentralized}. This data is crucial for the development of Artificial Intelligence (AI). However, the traditional paradigm of centralized AI model training, which aggregates all data on a central server, has become increasingly difficult to achieve due to growing concerns over data privacy and security.
Federated Learning (FL) has emerged as a promising paradigm for training machine learning models across distributed devices without sharing raw data~\cite{mcmahan2017communication}. Despite impressive theoretical and experimental advancements, FL still faces notable challenges~\cite{kairouz2021advances,wang2021field} in EC scenarios, such as healthcare~\cite{antunes2022federated}, recommendation systems~\cite{zhang2023fine}, and financial services~\cite{zheng2021federated}. A considerable challenge is the need for multiple communication rounds between several clients and a central server, which can be costly and intolerable due to the associated time and energy constraints~\cite{wu2022communication,xiong2023feddm}. 
Moreover, frequent communication poses a high risk of privacy attacks~\cite{lyu2022privacy, MiaoKLLLM24}, such as a man-in-the-middle attack~\cite{wang2020man} or the potential for reconstructing training data from gradients~\cite{yin2021see}. To address the communication and security challenges in conventional FL, the concept of one-shot FL has been introduced~\cite{guha2019one}, which aims to obtain an acceptable global model within a single communication round. 

\begin{figure}[t]
    \centering
    \includegraphics[width=0.9\linewidth]{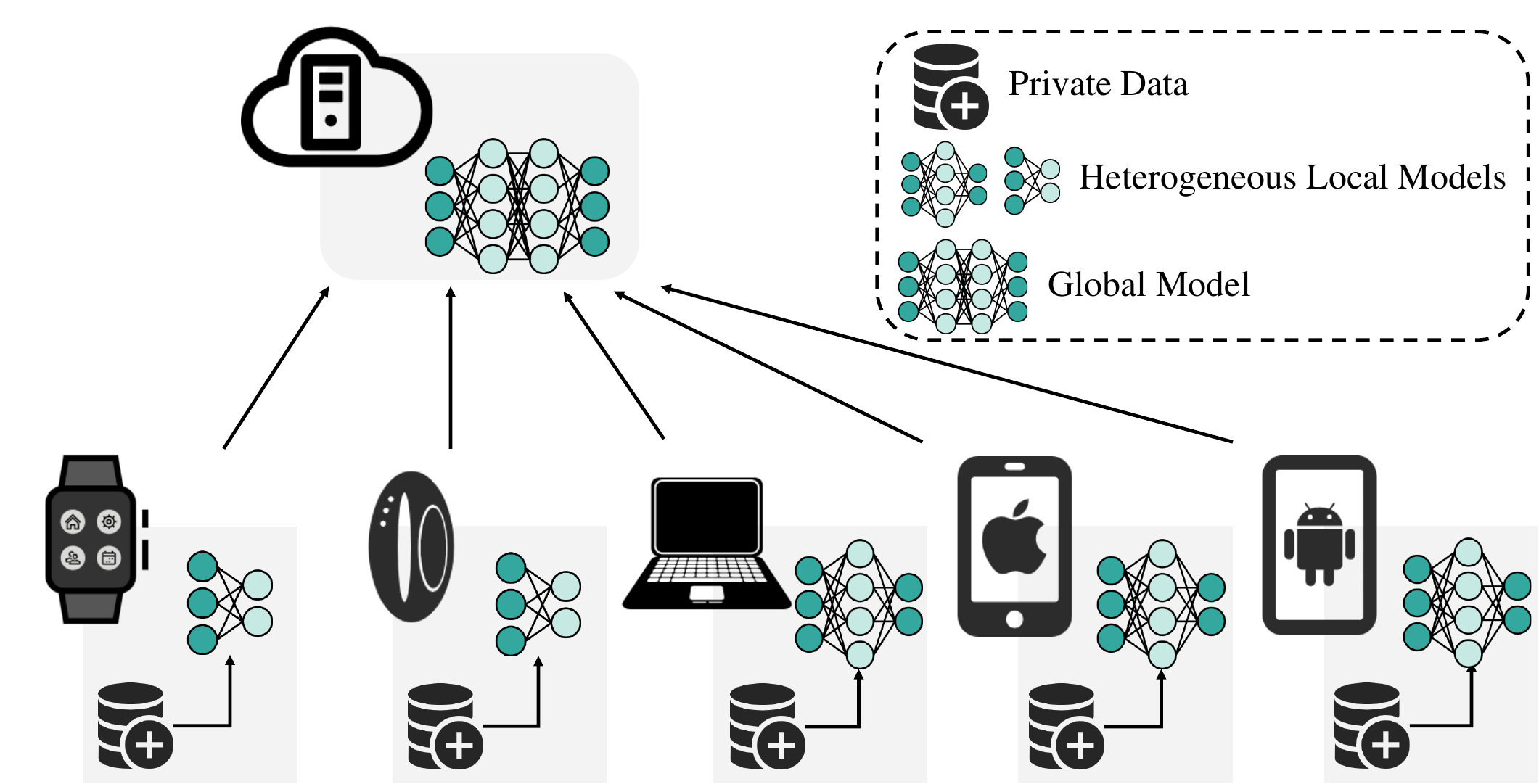}
    \caption{The heterogeneous one-shot Federated Learning (FL) framework. The clients with varying computing capabilities deploy heterogeneous local models. Each client communicates with the central server only once. After fully training their local models on private data, clients send these models to the server, where they are aggregated into a global model.}
    \label{figure_intro_framework}
\end{figure}

Another challenge in real-world FL scenarios for EC is the heterogeneity of client computing resources~\cite{zhang2024model}.
For example, a naive solution for developing a health monitoring model via FL involves training local models on each user's smartphone, smartwatch, or smart wristband, followed by the weighted averaging of their parameters. Smartphones typically have 4GB-16GB of RAM, while the Xiaomi Smart Wristband 8 offers 1.4MB of RAM, and the Amazfit Band 7 provides 8MB of RAM. This heterogeneity of client computing resources is common in EC applications of FL~\cite{wang2024achieving}. Existing FL frameworks typically require clients to deploy homogeneous local models. However, deploying uniformly small models prevents clients with abundant computing resources from fully utilizing their computational potential. Conversely, deploying uniformly large models excludes clients with limited computing resources from participating in the FL process, thus preventing the global model from acquiring knowledge from these resource-constrained clients. As a result, the deployment of homogeneous local models can degrade the performance of the final global model. A prevalent method for addressing computing heterogeneity is to deploy heterogeneous models on different clients, according to their respective computing capabilities~\cite{niu2024flrce}. Therefore, in FL scenarios with a mix of resource-sufficient clients and resource-constrained clients, as illustrated in Figure~\ref{figure_intro_framework}, studying model-heterogeneous one-shot FL becomes critical.

The existing one-shot FL methods~\cite{zhang2022dense,heinbaugh2023datafree} still face various challenges in model heterogeneity scenarios. 
DOSFL~\cite{zhou2020distilled} adopts dataset distillation~\cite{wang2018dataset} on each client. However, sending distilled data to the central server introduces additional communication costs, which goes against the original intention of addressing the communication bottleneck in one-shot FL. Some other methods~\cite{guha2019one, Li2020PracticalOF} leverage knowledge distillation~\cite{hinton2015distilling} to aggregate local models with auxiliary public data. However, the performance of these methods depends both on the size of the auxiliary data and on its domain similarity to the local data~\cite{stanton2021does}. Several data-free methods~\cite{zhang2022dense,heinbaugh2023datafree,dai2024enhancing} have been proposed to address these issues. DENSE~\cite{zhang2022dense} and Co-Boosting~\cite{dai2024enhancing} train an additional generator on the server side to generate auxiliary data that resembles the local data distribution. However, the efficacy of this generator heavily depends on the local model used for its training. If clients with limited computing resources deploy lightweight local models, whose performance is restricted, the generator's performance will suffer, thereby hindering the final global model from achieving high performance. FEDCVAE~\cite{heinbaugh2023datafree} deploys generative models on all participating clients, with the global model on the server being trained from scratch using the synthetic samples generated by these generators. Nevertheless, using synthetic samples to convey local data information is less effective than directly using local model parameters. A more detailed explanation can be found in Section~\ref{section_motivation}.


In this paper, we propose a novel method, FedMHO, to address the challenge of \textbf{M}odel-\textbf{H}eterogeneous \textbf{O}ne-shot \textbf{Fed}erated Learning. Our method involves deploying deep classification models on resource-sufficient clients while utilizing lightweight generative models on resource-constrained clients. In this paper, we adopt Conditional Variational Autoencoders (CVAE)~\cite{sohn2015learning} as the generative models to validate the effectiveness of our methods. More complex generators can be employed for more complex tasks. During global model training, FedMHO introduces a two-stage process encompassing data generation and knowledge fusion. In the data generation stage, the decoders received from clients generate synthetic samples based on local label distribution. To improve the fidelity of the synthetic samples, we employ an unsupervised data optimization solution. In the subsequent knowledge fusion stage, the global model is initialized by the average parameters of the local classification models and then updated based on the synthetic samples generated in the data generation stage. Furthermore, during the global model training, we propose two solutions named FedMHO-MD and FedMHO-SD to prevent the forgetting of knowledge from classification models. In FedMHO-MD, the local classification models act as multiple teacher models to distill the global model. FedMHO-SD involves self-distillation, where the initialized global model acts as a teacher to guide the training of the global model. 
Our main contributions are summarized as follows.
\begin{itemize}
\item We propose FedMHO, a one-shot FL framework designed for heterogeneous clients. This framework deploys classification models on computing resource-sufficient clients and lightweight generative models on computing resource-constrained clients.
\item We propose an unsupervised data processing solution to optimize the quality of synthetic samples, consequently improving the performance of the final global model.
\item To address the knowledge-forgetting problem during the training of the global model, we propose two effective strategies: FedMHO-MD and FedMHO-SD.
\item Compared to the optimal baseline, the average accuracies of FedMHO, FedMHO-MD, and FedMHO-SD are improved by 5.17\%, 8.35\%, and 8.25\%, respectively, demonstrating the effectiveness of the proposed methods.
\end{itemize}

\section{Related Work}


\subsection{One-shot Federated Learning}

The one-shot FL methods aim to maximize the aggregation of data information from clients into the global model within a single communication round.~\cite{guha2019one} originally proposes one-shot FL and introduces two aggregation methods: the first one ensembles the local models via bagging strategies, and the second one uses ensemble distillation with auxiliary public data. FedOV~\cite{heinbaugh2023datafree} uses open-set voting~\cite{neal2018open,zhou2021learning} to solve the problem of label skewness when bagging local models. However, bagging-like strategies can introduce additional computational overhead during inference on the client side. FedKT~\cite{Li2020PracticalOF} uses knowledge distillation for aggregating local models into a global model based on auxiliary public data, which is similar to the first of the methods proposed in~\cite{guha2019one}. 

To achieve data-free one-shot FL, DOSFL~\cite{zhou2020distilled} employs dataset distillation~\cite{wang2018dataset}. Each client distills its local data into a small dataset and sends it to the server for global model training. DENSE~\cite{zhang2022dense} trains a generator on the server side, and then uses local models to distill the global model based on synthetic samples produced by the generator. Based on DENSE, Co-Boosting~\cite{dai2024enhancing} further improves the global model performance by optimizing the data and improving the integration. Two-stage methods like DENSE and Co-Boosting, which first train the classifier and then train the generator, cause information loss twice in the two training stages. In order to reduce the double information loss, FedSD2C~\cite{zhang2024one} synthesizes samples directly from local data and proposes to share synthetic samples instead of inconsistent local models to solve the problem of data heterogeneity. However, FedSD2C is only applicable to large-scale datasets with high resolution. \cite{heinbaugh2023datafree} propose FEDCVAE-ENS and FEDCVAE-KD, which deploy generative models locally and generate data based on local label distributions on the server to train the global model. The limitation of the above data-free methods is that they do not take into account the fact that the application scenarios of one-shot FL are usually accompanied by model heterogeneity.

\subsection{Model-Heterogeneous Federated Learning}

To achieve model heterogeneity in FL, some methods extract heterogeneous sub-models from the global model to use as local models. For example, Federated Dropout~\cite{caldas2018expanding} randomly extracts sub-models using Dropout~\cite{srivastava2014dropout}. 
HeteroFL~\cite{diao2020heterofl}, FjORD~\cite{horvath2021fjord}, and FedDSE~\cite{wang2024feddse} extract static sub-models from the global model, while FedRolex~\cite{alam2022fedrolex} and Split-Mix~\cite{hong2022efficient} extract dynamic sub-models. These partial training-based methods require each local model to be a sub-model of the global server model, preventing their deployment in FL scenarios involving completely different model structures.

Knowledge distillation-based methods do not have the limitations of PT-based methods and are more suitable for FL scenarios with diverse models. FedMD~\cite{li2019fedmd} uses transfer learning~\cite{liang2020we} to pretrain local models on large-scale public datasets, then fine-tunes them using local datasets. FedGKT~\cite{he2020group} and FedDKC~\cite{wu2024exploring} use Split Learning~\cite{gupta2018distributed} with the global model serving as a downstream model of local models. Both methods rely on sending label information of local data from clients to the server, exposing the risk of privacy leakage. Methods such as FML~\cite{shen2023federated}, PervasiveFL~\cite{xia2022pervasivefl}, and FedKD~\cite{wu2022communication} assign both a public and a private model to each client. The public model is homogeneous across all clients, while the private model is heterogeneous. By introducing Mutual Learning~\cite{zhang2018deep}, the public and private model can interact with each other locally. However, maintaining two models on each client introduces additional computation and storage overhead.
FedDF~\cite{lin2020ensemble}, DS-FL~\cite{itahara2021distillation}, and Fed-ET~\cite{ijcai2022p399} utilize unlabeled auxiliary data to transfer knowledge from local models to the global model. However, The differences in distribution between auxiliary and local training data can affect the accuracy of the global model~\cite{stanton2021does}.

To address the limitations of using public datasets as auxiliary data, some studies have introduced data-free knowledge distillation techniques~\cite{chen2019data, do2022momentum}. For example, FedFTG~\cite{zhang2022fine} trained a pseudo data generator by fitting the input space of a local model. The generated pseudo data is then integrated into the subsequent distillation process. DFRD~\cite{wang2024dfrd} further improves the training of the pseudo data generator, which can generate synthetic samples more similar to the distribution of local training data, thereby improving the accuracy of the global model. However, the effectiveness of these methods is highly dependent on the quality of the local model. If the performance of the local model is poor, the generated pseudo data may also be of low quality, limiting the accuracy of the global model.



\section{Motivation}
\label{section_motivation}
Current FL algorithms fail to adequately address the challenge of one-shot FL with heterogeneous models. 
Model-heterogeneous FL algorithms can be broadly categorized into partial training-based methods and knowledge distillation-based methods. However, \textbf{partial training-based methods are essentially not applicable to one-shot scenarios, and knowledge distillation-based methods introduce misleading information that misleads the global model}.

Partial training-based methods assign heterogeneous sub-models to clients, either by randomly dropping out parameters from the global model or extracting them from the global model based on specific rules. Since one-shot FL involves only a single round of communication and the local model parameters are typically fractions (e.g., $\frac{1}{2}$, $\frac{1}{4}$, $\frac{1}{8}$) of the global model, most of the global model parameters are updated by local data from only a subset of clients. This constraint leads to poor performance of the aggregated global model, and no existing work successfully combines partial training with one-shot FL. 

\begin{figure}[t]
    \centering
    \includegraphics[width=0.95\linewidth]{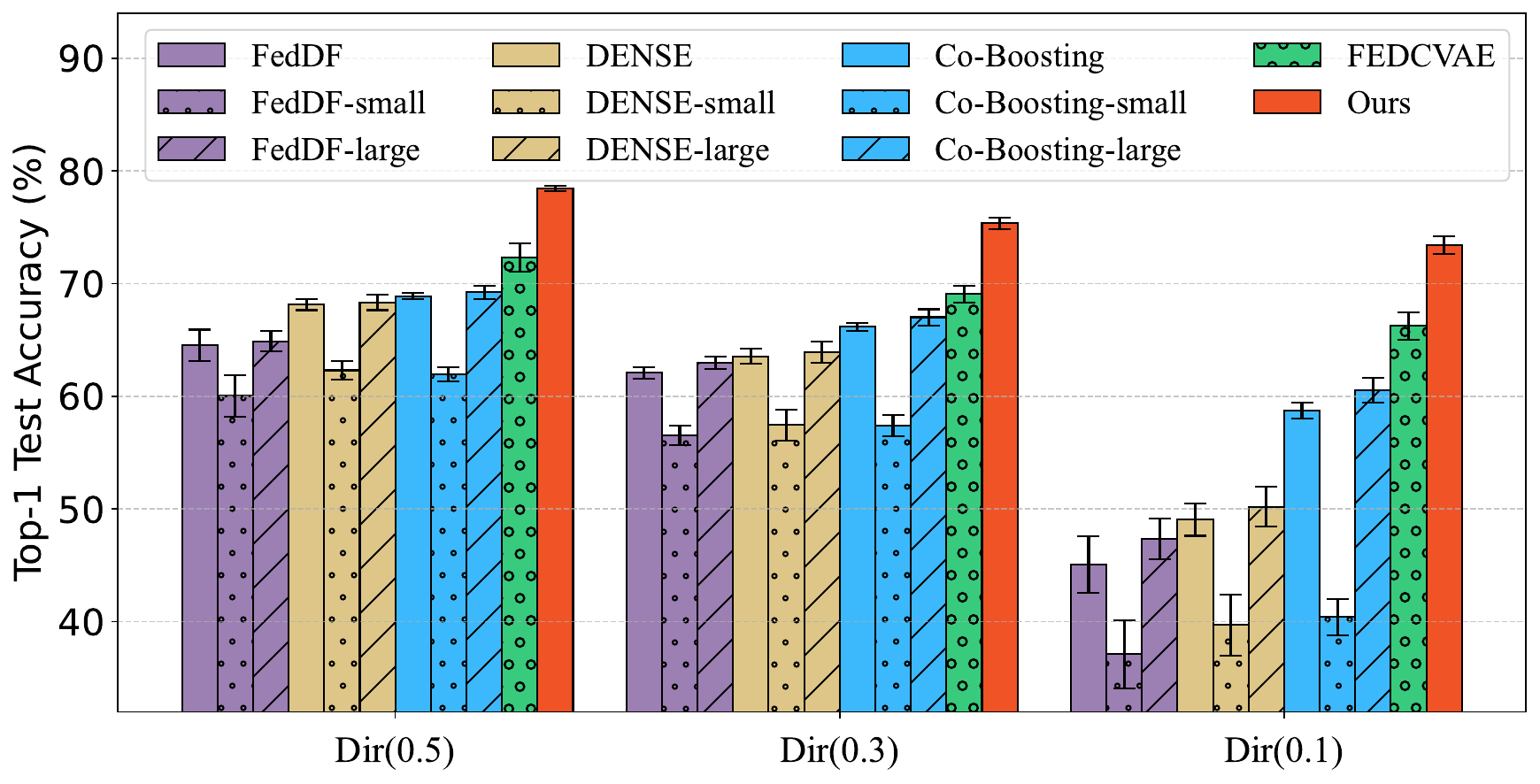}
    \caption{Top-1 test accuracy (\%)  on the EMNIST dataset. `small' or `large' refers to the aggregation of only lightweight small local models or only deep large local models, respectively.
}
    \label{figure_motivation}
\end{figure}

Knowledge distillation-based methods employ the local models as multiple teachers and utilize an auxiliary dataset to train the global model via knowledge distillation, which proves effective in one-shot scenarios~\cite{salehkaleybar2021one}.
This auxiliary dataset can be sourced from a public dataset, such as FedDF, or by an additional trained generator like DENSE or Co-Boosting. However, the lightweight local models of FedDF, DENSE, or Co-Boosting often underperform and provide inaccurate logits (i.e., soft labels) during training of the global model or generator~\cite{chen2021distilling}, which subsequently degrade the performance of the final global model. Figure~\ref{figure_motivation} presents the experimental results on the EMNIST dataset. The experimental setup is the same as in Section~\ref{section_experiments}. Notably, when only the deep large models are used, the global model's performance slightly improves compared to using all local models, even though data from the lightweight small models is not aggregated. This indicates that lightweight small models provide a negative gain.

FEDCVAE deploys generative models on all clients to generate synthetic samples for training the global model. Figure~\ref{figure_motivation} demonstrates that FEDCVAE has a significantly better performance compared to FedDF, DENSE and Co-Boosting. This improvement arises because, while lightweight generative models may produce low-quality samples, such as blurred contours (refer to Figure~\ref{figure_visualization_synthetic_samples} in the Appendix), they do not introduce erroneous logits like poorly performing classification models do.

Nonetheless, synthetic samples cannot fully match the quality of the raw data. To maximize the use of real data and minimize erroneous information from low-performance models, we propose deploying deep classification models on clients with sufficient computing resources and lightweight generative models on clients with limited computing resources. During aggregation, the classification models directly average the parameters to initialize the global model, and then the global model is further trained by the synthesized samples. The effectiveness of our proposed method can also be demonstrated in Figure~\ref{figure_visualization_synthetic_samples}.

\begin{figure*}[t]
    \centering \includegraphics[width=0.9\linewidth]{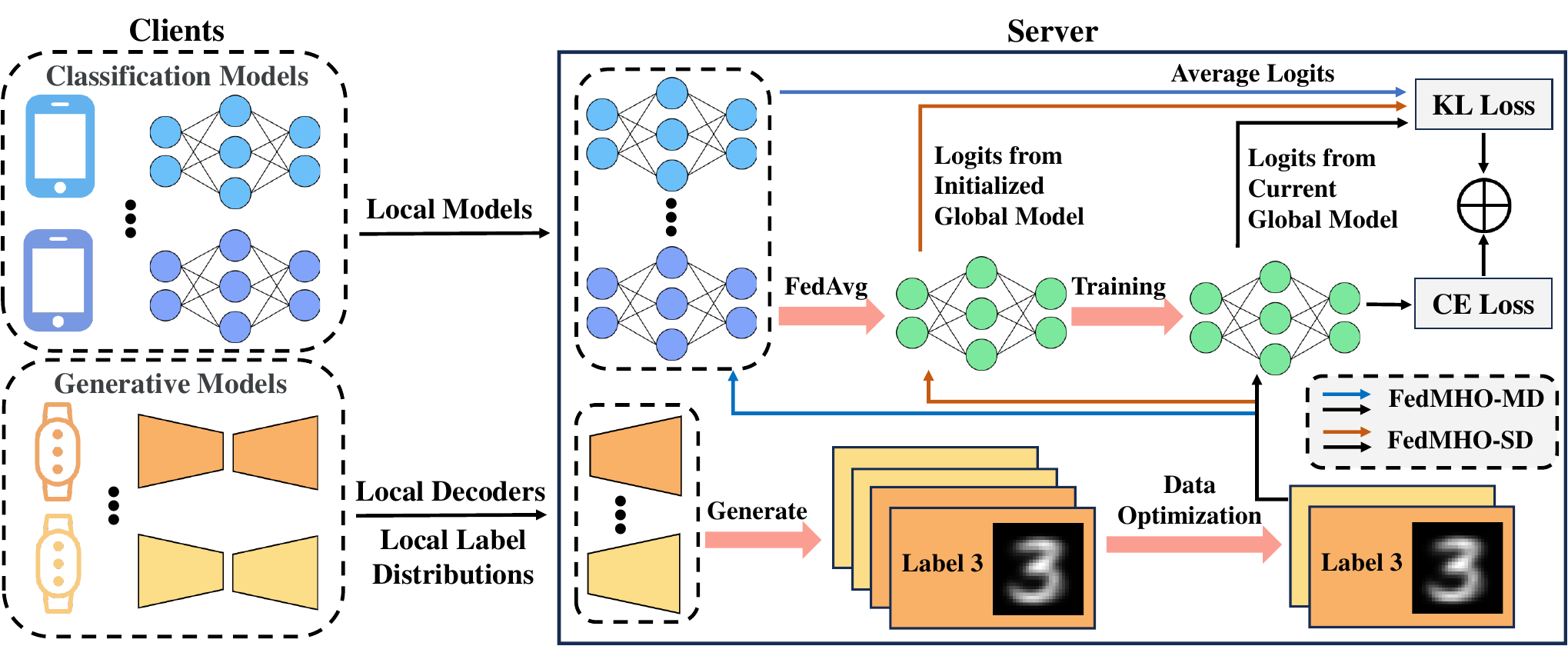}
    \caption{An overview of FedMHO. Resource-sufficient clients train deep classification models, and resource-constrained clients train lightweight generative models. The global model is initialized by averaging local classification models. Synthetic samples are generated to train the global model. To solve the knowledge-forgetting problem, FedMHO-MD employs classification models as teachers to distill the global model, while FedMHO-SD utilizes the initialized global model as a teacher to distill the global model.}
    \label{figure_framework}
\end{figure*}

\section{Methodology}


\subsection{Preliminaries}

\subsubsection{Conditional Variational Autoencoder}
CVAE is a variation of Variational Autoencoder (VAE)~\cite{kingma2013auto} that incorporates conditional information into VAE. Specifically, a VAE learns a latent space representation of input data through an encoder-decoder architecture. In CVAE, this latent space is conditioned on additional information, such as class labels, enabling controlled generation based on specific conditions. Formally, let $\mathbf{x}$ denote the input data, $\mathbf{c}$ denote the conditional information, $\phi$ denotes the encoder network, and $\theta$ denotes the decoder network, respectively. A CVAE uses $\phi$ to model the approximate posterior distribution $q_{\phi}(\mathbf{z}|\mathbf{x},\mathbf{c})$ over latent variables $\mathbf{z}$, and uses $\theta$ to model the generative distribution $p_{\theta}(\mathbf{x}|\mathbf{z},\mathbf{c})$ for reconstructing the input data.




\subsubsection{Knowledge Distillation}
Knowledge distillation aims to minimize the discrepancy between logits output from the teacher model and the student model on the same data. The discrepancy can be measured by the Kullback–Leibler divergence~\cite{kullback1951information} $\textbf{KL}[\cdot]$ as $\min\limits_{W_S}\textbf{KL}_{x\in D}[h(W_T;x)||h(W_S;x)]$, where $W_T$ denotes teacher model, $W_S$ denotes student model, $D$ denotes the data used for knowledge distillation, and $h(*;x)$ denotes the output logits of input data $x$.



\subsubsection{Problem Definition}
We consider a one-shot FL setup with an $N_c$-class classification task, where $K$ clients are connected to a central server. We define the set of clients as $\{K\} = \{ \{K_C\}, \{K_G\} \}$, where $\{K_C\}$ denotes the clients that develop classification models, and $\{K_G\}$ denotes clients that develop generative models. Each client $k \in \{K\}$ holds a local model $w_k$ and local training data $\mathcal{D}_k$. Specifically, a computing resource-sufficient client $k$ holds a local classification model $\{ w_k\:|\:k \in \{K_C\} \}$, while a computing resource-constrained client $k$ holds a lightweight generative model $\{ w_k\:|\:k \in \{K_G\} \}$ with an encoder $w_k^{\phi}$ and a decode $w_k^{\theta}$. We aim to join resource-constrained clients in FL training and train a well-performing global model $w$ with resource-sufficient clients.

\begin{algorithm}[t]
    \caption{The proposed FedMHOs Algorithm}
    \label{algorithm_FedMHO}
    \begin{algorithmic}[1] 
        \STATE \textbf{Input}: Number of categories $N_c$, local model $w_k$, local training epoch $E_k$, global training epoch $E_g$, the set of local classification model $\{K_C\}$, the set of generative model decoder $\{K_G\}$.\\
        \STATE \textbf{Output}: The final global model $w$.\\
        
        \STATE /* Client Side */
        \FOR{each $\mathcal{E} = 1 \cdots E_k$}
            \STATE Train $w_k$ on $\mathcal{D}_k$ with \eqref{equation_ce} ($k \in \{K_C\}$) or \eqref{equation_train_cvae} ($k \in \{K_G\}$).
            
        \ENDFOR

        \STATE \textbf{return} updated $w_k$ ($k \in \{K_C\}$) or updated $w_k^{\theta}$ ($k \in \{K_G\}$).
        
        
        \STATE /* Server Side */
        \STATE Initialize global model $w_0$ with \eqref{equation_fedavg}.

        \STATE Initialize an empty set for generate samples $\mathcal{D}_s$.
        
        \FOR{$n_c = 1 \cdots N_c$}
            \STATE Generate samples belonging to category $n_c$ using decoder $ \{ w_k^{\theta}\:|\: k \in \{K_G\} \}$.
            \STATE Optimize synthetic samples via K-means and add to $\mathcal{D}_s$.
        \ENDFOR
        \FOR{$\mathcal{E} = 1 \cdots E_g$}
            \STATE \textbf{case} FedMHO Alg.: 
            \STATE \quad\quad Train global model $w$ on $\mathcal{D}_s$ with Eq.~\eqref{equation_fedmho}.
            \STATE \textbf{case} FedMHO-MD or FedMHO-SD Alg.: 
            \STATE \quad \quad Train global model $w$ on $\mathcal{D}_s$ with Eq.~\eqref{equation_boost_global_model}.
        \ENDFOR
        \STATE \textbf{return} final global model $w$.

    \end{algorithmic}
\end{algorithm}

\subsection{Overall Algorithm}

Figure~\ref{figure_framework} illustrates the overall framework of FedMHO. Specifically, resource-sufficient clients deploy deep classification models, and resource-constrained clients deploy lightweight CVAEs. Each client fully trains its respective local model during the local training phase. Afterwards, clients with classification models send their complete models to the central server, and clients utilizing CVAEs send the CVAE decoders and the local label distributions to the central server. The server-side training process consists of data generation and knowledge fusion. In the data generation stage, each local CVAE decoder generates synthetic samples based on its local label distribution. These samples are then refined through an unsupervised quality enhancement process. In the knowledge fusion stage, the local classification models initialize the global model by aggregating their parameters. The global model is then updated with the synthetic samples. To mitigate knowledge-forgetting during the global model training, we employ FedMHO-MD and FedMHO-SD. The specific training details are provided below. The complete algorithmic representation of our methods is provided in Algorithm~\ref{algorithm_FedMHO}.

\subsubsection{Local Training}
The clients train their local models on their respective local data for $E_k$ epochs. During each training epoch, each local model weight $w_k$ is updated as
\begin{equation}
    w_k := w_k-\eta \cdot \nabla \mathcal L_k(w_k;b_l),
\end{equation}
where $\eta$ denotes the learning rate, $b_l$ denotes the mini-batch from local data $\mathcal{D}_k$, $L_k$ denotes the training loss function of local model $k$, and $\nabla \mathcal L_k(\cdot)$ denotes the partial derivative of $ \mathcal L_k(\cdot) $ with respect to its parameter $w_k$. For the classification models, $\mathcal{L}_k(w_k; b_l)$ is defined as the cross-entropy loss
\begin{equation}
    \label{equation_ce}
    \mathcal{L}_k(w_k; b_l) = - \sum\nolimits_{i=1}^{|b_l|} y_i \log(\hat{y}_i),
\end{equation}
where $|b_l|$ denotes the size of the mini-batch $b_l$, $y_i$ and $\hat{y}_i$ represent the ground truth and  predicted probability of sample $i$, respectively. For the CVAE models, $\mathcal{L}_k(w_k; b_l)$ is the sum of the reconstruction loss $\mathcal{L}_{\text{recon}}$ and the KL divergence loss $\mathcal{L}_{\text{KL}}(w_k; b_l)$, defined as
\begin{equation}
    \label{equation_train_cvae}
    \mathcal{L}_k(w_k; b_l) = \mathcal{L}_{\text{recon}}(w_k; b_l) + \mathcal{L}_{\text{KL}}(w_k; b_l),
\end{equation}
with 
\begin{align*}
    \mathcal{L}_{\text{recon}}(w_k; b_l) = \frac{1}{|b_l|} \sum\nolimits_{i=1}^{|b_l|} \mathbb{E}_{q_{w_k^{\phi}}(\mathbf{z}|x_i, \mathbf{c}_i)}\left[\log p_{w_k^{\theta}}(x_i|\mathbf{z}, \mathbf{c}_i)\right], \\
    \mathcal{L}_{\text{KL}}(w_k; b_l) = -\frac{1}{|b_l|} \sum\nolimits_{i=1}^{|b_l|} \text{KL}\left(q_{w_k^{\phi}}(\mathbf{z}|x_i, \mathbf{c}_i) || p(\mathbf{z}|\mathbf{c}_i)\right),
\end{align*}
where $x_i$ denotes sample $i$ in mini-batch $b_l$, $\mathbf{c}_i$ denotes the conditional information of sample $i$, that is, the ground truth of $x_i$. The term $q_{w_k^{\phi}}(\mathbf{z}|x_i, \mathbf{c}_i)$ refers to the approximate posterior distribution of local model $k$, and the term $p_{w_k^{\theta}}(x_i|\mathbf{z}, \mathbf{c}_i)$ refers to the generative distribution of local model $k$. The term $p(\mathbf{z}|\mathbf{c}_i)$ refers to the prior distribution, which is typically assumed to follow a standard Gaussian distribution $\mathbf{z} \sim \mathcal{N}(0,1)$. The reconstruction loss measures the difference between the generated samples and the input samples, and the KL divergence loss ensures that the distribution of the latent space of the synthetic samples approaches the prior distribution $\mathbf{z}$.

\subsubsection{Data Generation}

During the data generation stage, the server utilizes the received local decoders and local label distributions to generate synthetic samples $\mathcal{D}_s$. Specifically, for each decoder $\theta_k \in \{K_G\}$, we sample $x \in \mathcal{D}_s$ according to client $k$'s local label distribution, formally expressed as $x_i \sim p_{\theta_k} \left( x_i |\mathbf{z}, \mathbf{c}_i \right)$,
where $\mathbf{z} \sim \mathcal{N}(0,1)$, $\mathbf{c}_i$ is the label of $x_i$, and $p_{\theta_k} \left( x |\mathbf{z}, \mathbf{c}_i \right)$ represents the conditional probability distribution defined by the client $k$'s decoder $\theta_k$. The synthetic samples are subsequently used in knowledge fusion sessions. As the generators are lightweight and the local data partitions often exhibit varying degrees of Non-IID, the generated samples may incorporate a certain level of noise. To enhance the quality of these samples, we introduce an unsupervised solution, which is detailed in Section~\ref{section_unsupervised_data_optimization}.

\subsubsection{Knowledge Fusion}
\label{section_knowledge_fusion}

During the knowledge fusion stage, the global model is initialized by the local classification models as
\begin{equation}
    \label{equation_fedavg}
    w_{0}=\frac{1}{|\{K_C\}|}\sum\nolimits_{k \in \{K_C\}} w_k,
\end{equation}
where $\{K_C\}$ represents the set of local classification models, and $|\{K_C\}|$ denotes the number of elements in $\{K_C\}$. This initialization enables the global model to incorporate knowledge from these local models. Compared with random parameter initialization, this informed initialization allows the global model to achieve superior performance in fewer training rounds. We specifically use $w_0$ to denote the initial state of the global model, while $w$ refers to its later states. To additionally obtain knowledge from the clients deploying generative models, we train the global model using the synthetic samples generated during the data generation stage. The training loss function $\mathcal{L}_\text{CE}(w;b_s)$ of the global model is defined as
\begin{equation}
    \label{equation_fedmho}
    \mathcal{L}_\text{CE}(w; b_s) = - \sum_{i=1}^{|b_s|} y_i \log(\hat{y}_i),
\end{equation}
where $b_s$ denotes the mini-batch of synthetic samples from $\mathcal{D}_s$.

\begin{figure}[tp]
    \centering
    \includegraphics[width=0.98\linewidth]{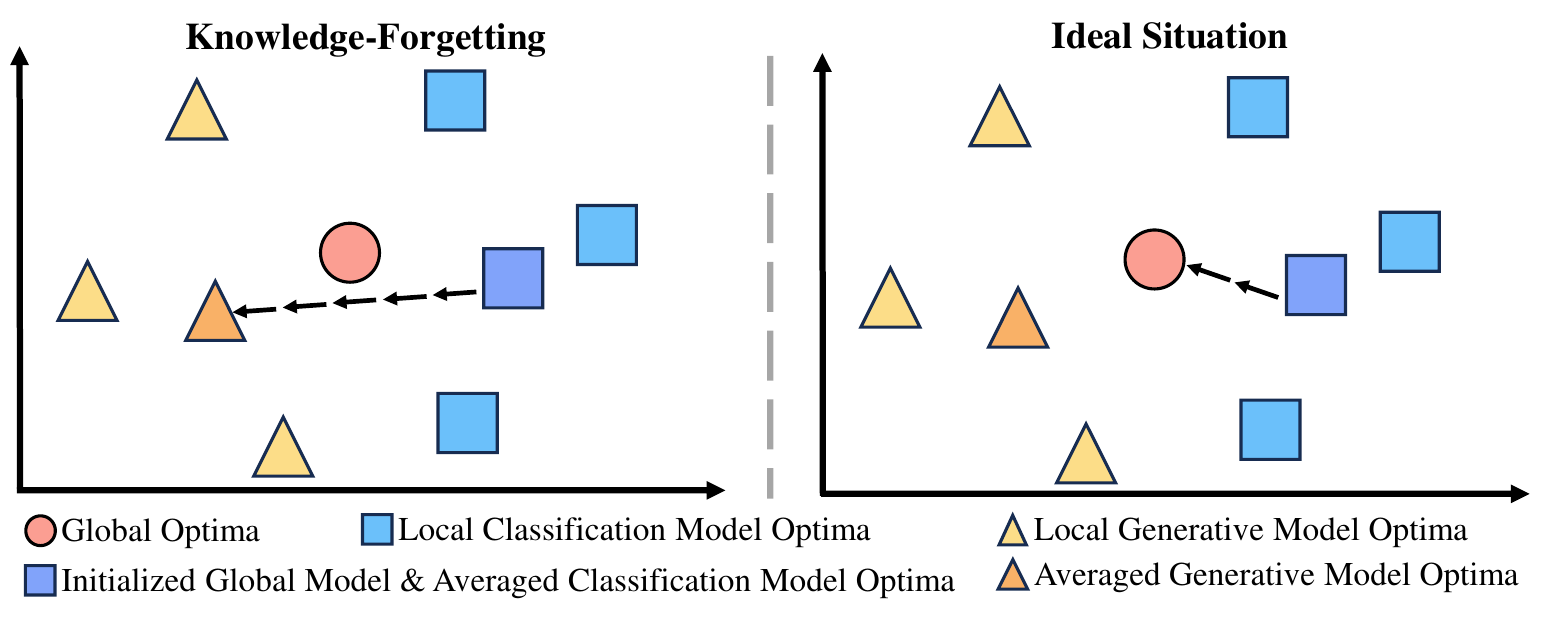}
    \caption{A toy example of the knowledge-forgetting problem.}
    \label{figure_toy_example_knowledge_forgetting}
\end{figure}

\subsection{The Knowledge-Forgetting Problem}
During the fusion of the clients' knowledge from the generative models, the global model may forget the knowledge learned from the classification models, as depicted in Figure~\ref{figure_toy_example_knowledge_forgetting}. We define this phenomenon as the knowledge-forgetting problem. We propose two methods to alleviate this problem: FedMHO-MD and FedMHO-SD. In FedMHO-MD, we use local classification models as multiple teacher models to distill their knowledge into the global model. The corresponding loss function $\mathcal{L}_{\text{KL}}(w;b_s)$ is defined as
\begin{equation}
    \mathcal{L}_{\text{KL}}(w; b_s) = \text{KL} \left[ \left(\frac{1}{|\{K_C\}|} \sum\limits_{k\in \{K_C\}}h_k(w_k,b_s) \right) || h(w,b_s) \right],
\end{equation}
where $h_k(w_k,b_s)$ and $h(w,b_s)$ denote the output logits of samples in the mini-batch $b_s$ through local model $w_k$ and global model $w$, respectively. In FedMHO-SD, we use the initialized global model $w_0$ as the teacher model to self-distill the global model. The corresponding loss function $\mathcal{L}_{\text{KL}}(w;b_s)$ is defined as
\begin{equation}
    \mathcal{L}_{\text{KL}}(w; b_s) = \text{KL}\left[h_k(w_0,b_s) || h(w,b_s)\right].
\end{equation}
where $h_k(w_0,b_s)$ denotes the output logits of samples in the mini-batch $b_s$ through the initialized global model $w_0$. Overall, the complete loss function $\mathcal{L}_g$ for training the global model is defined as:
\begin{equation}
    \label{equation_boost_global_model}
    \mathcal{L}_g(w;b_s) = \lambda \mathcal{L}_{\text{CE}}(w;b_s) + (1 - \lambda) \mathcal{L}_{\text{KL}}(w;b_s),
\end{equation}
where $\lambda \in [0, 1]$ denotes the trade-off parameter between $\mathcal{L}_{\text{CE}}$ and $\mathcal{L}_{\text{KL}}$. In our experiments, we observe the values of $\mathcal{L}_{\text{CE}}$ and $\mathcal{L}_{\text{KL}}$ are of similar magnitude, thus, we set $\lambda=0.5$ by default.

\begin{algorithm}[t]
    \caption{The Unsupervised Data Optimization Algorithm}
    \label{algorithm_data_clean}

    \begin{algorithmic}[1] 
        \STATE \textbf{Input}: Number of categories $N_c$, synthetic samples $\mathcal{D}_s^c$, the ratio of the number of remaining samples to the number of original samples $\mathcal{R}_{th}$, feature dimension $F$.\\
        \STATE \textbf{Output}: Optimized synthetic sample $\mathcal{D}_s$. \\
        
        \FOR{$n_c = 1 \cdots N_c$ \textbf{in parallel}}
            \STATE Select the synthetic samples with the label $n_c$ to form the dataset $\mathcal{D}_s^c$.
            \STATE Cluster $\mathcal{D}_s^c$ into one category using the K-means algorithm and get the cluster center $C_c$.
            \FOR{\textbf{each} sample $i \in \mathcal{D}_s^c$ \textbf{in parallel}}
                \STATE {} /* Calculate the Euclidean Distance */ 
                \STATE $Dis_i = (\sum_{f \in F} (i^f - C_c^f)^2)^{1/2}$. 
            \ENDFOR
            \STATE Filter out the $\left(1-\mathcal{R}_{th}\right) \times |\mathcal{D}_s^c|$ samples with the largest $Dis_i$ and update $\mathcal{D}_s$.
        \ENDFOR

        \STATE \textbf{return} Optimized $\mathcal{D}_s$.
    \end{algorithmic}
\end{algorithm}

\subsection{Unsupervised Data Optimization}
\label{section_unsupervised_data_optimization}

We observe that a lightweight CVAE occasionally generates samples that exhibit label confusion. For example, when the decoder receives an input conditioned on label 2 for generating synthetic samples, the model might produce samples with labels 1 or 3 instead. While such instances of mislabeling are infrequent, they introduce noise into the training data of the global model. To enhance the quality of synthetic samples and improve the global model, we propose an unsupervised data optimization strategy based on the K-means algorithm~\cite{lloyd1982least}.

Specifically, for synthetic samples corresponding to a specific class $n_c \in N_c$, we use the flattened pixel values as features and apply K-means clustering to group these samples into one cluster. The feature dimension $F$ is determined by the number of pixels in the image. For example, for an image of size $10\times 10$ pixels, the feature dimension $F$ is 100. After clustering, we obtain the cluster center $C_c$ corresponding to each category $n_c$. Subsequently, we selectively retain the samples closest to each $C_c$. By default, we set the ratio of the number of remaining samples to the number of original samples, denoted as $\mathcal{R}_{th}$, to 80\%. Experimental results for different values of $\mathcal{R}_{th}$ are shown in Section~\ref{section_ratio_of_data_optimization}. A detailed algorithmic description is provided in Algorithm~\ref{algorithm_data_clean}.

\section{Experiments}
\label{section_experiments}


\begin{table}[t]
    \normalsize
    \caption{Diverse local models used in different methods.}
    \label{table_local_model}
    \centering
    \addtolength{\tabcolsep}{-1.5pt}
    \begin{tabular}{c|ccc}
    \toprule
        Model & \makecell[c]{FedAvg, FedDF, \\DENSE, Co-Boosting} & FEDCVAE & FedMHOs\\
      
        \cmidrule(lr){1-4}
        Large & VGG-9 & CVAE-large & VGG-9 \\
        Small & CNN & CVAE-small & CVAE-small \\
        
        \bottomrule
    \end{tabular}
\end{table}

\begin{table}[t]
    \normalsize
    \caption{FLOPs of various models with batch size set to 1. }
    \label{table_flops}
    \centering
    \begin{tabular}{c|ccc}
    \toprule
        Model & MNIST \& Fashion & SVHN & EMNIST \\
      
        \cmidrule(lr){1-4}
        VGG-9 & 126.47M & 145.48M & 126.47M\\
        CNN & 467.23K & 4.00M & 3.96M\\
        CVAE-large & 152.67M & 201.76M & 152.67M\\
        CVAE-small & 408.06K & 2.08M & 1.06M\\

        \bottomrule
    \end{tabular}
\end{table}

\begin{table}[t]
    \normalsize
    \caption{Detailed hyperparameter settings utilized in experiments. `LR' is short for `Learning Rate'.}
    \label{table_hyperparameter}
    \addtolength{\tabcolsep}{-4.7pt}
    \centering
    \begin{tabular}{c|cccc}
    \toprule
        Hyperparameter & MNIST & Fashion & SVHN & EMNIST\\
      
        \cmidrule(lr){1-5}
        Batch Size & \multicolumn{4}{c}{64} \\
        Local Epoch (VGG-9 \& CNN) & \multicolumn{4}{c}{200} \\
        Local Epoch (CVAE) & 30 & 40 & 40 & 50\\
        Local Optimizer (VGG-9) & \multicolumn{4}{c}{SGD} \\
        Local LR (VGG-9) & 5$e$-3 & 5$e$-3 & 5$e$-3 & 1$e$-3 \\
        Momentum (VGG-9) & \multicolumn{4}{c}{0.9} \\
        Local optimizer (CVAE) & \multicolumn{4}{c}{Adam} \\
        Local LR (CVAE) & 5$e$-2 & 5$e$-2 & 1$e$-3 & 3$e$-3 \\
        Global Epoch & 10 & 20 & 20 & 30 \\
        Global Optimizer & \multicolumn{4}{c}{Adam} \\
        Global LR & 1$e$-5 & 5$e$-4 & 5$e$-5 & 5$e$-5 \\
        Number of Synthetic Data  & 6,000 & 6,000& 6,000 & 12,000 \\

        \bottomrule
    \end{tabular}
\end{table}

\begin{table*}[t]
    \normalsize
    \caption{Top-1 test accuracy (\%) comparison under various datasets and data partitions. \textbf{Bold} denotes the best result, and \underline{underline} denotes the second best result.}
    \label{table_performance_comparison}
    \centering
    \addtolength{\tabcolsep}{-3.5pt}
    \begin{tabular}{c|c|cccccccc}
    \toprule
        Dataset & Partition & FedAvg & FedDF & DENSE & Co-Boosting & FEDCVAE & FedMHO & FedMHO-MD & FedMHO-SD \\
        \cmidrule(lr){1-10}
        \multirow{3}{*}{MNIST} 
        & $Dir(0.5)$ & 85.91$\pm$0.57 & 88.73$\pm$1.29 & 90.51$\pm$2.39 & 91.42$\pm$1.44 & 92.61$\pm$0.25 & 93.45$\pm$0.65 & \textbf{95.71$\pm$0.68} & \underline{95.67$\pm$0.47}\\
        & $Dir(0.3)$ & 74.57$\pm$0.73 & 78.46$\pm$0.91 & 86.12$\pm$1.01 & 89.27$\pm$0.48 & 91.37$\pm$0.65 & 92.43$\pm$0.42 & \underline{93.98$\pm$0.69} & \textbf{94.25$\pm$0.18}\\
        & $Dir(0.1)$ & 49.12$\pm$2.99 & 62.58$\pm$4.50 & 73.76$\pm$1.72 & 81.30$\pm$0.54 & 89.73$\pm$0.40 & 88.48$\pm$0.99 & \underline{91.22$\pm$0.56} & \textbf{91.55$\pm$0.27}\\

        \cmidrule(lr){1-10}
        \multirow{3}{*}{Fashion}
        & $Dir(0.5)$ & 60.24$\pm$1.47 & 65.66$\pm$1.36 & 72.29$\pm$2.63 & 72.98$\pm$1.61 & 69.11$\pm$1.35 & 75.30$\pm$0.62 & \underline{77.27$\pm$0.48} & \textbf{77.73$\pm$0.59}\\
        & $Dir(0.3)$ & 47.36$\pm$0.21 & 59.21$\pm$2.20 & 68.14$\pm$0.97 & 69.50$\pm$0.64 & 68.39$\pm$0.49 & 72.36$\pm$0.92 & \underline{75.69$\pm$0.93} & \textbf{76.45$\pm$0.53}\\
        & $Dir(0.1)$ & 30.37$\pm$0.79 & 42.21$\pm$2.80 & 56.12$\pm$3.94 & 61.73$\pm$1.36 & 64.01$\pm$2.47 & 62.14$\pm$1.45 & \textbf{71.04$\pm$0.50} & \underline{69.65$\pm$0.99} \\

        \cmidrule(lr){1-10}
        \multirow{3}{*}{SVHN}
        & $Dir(0.5)$ & 60.73$\pm$1.91 & 72.23$\pm$1.32 & 77.61$\pm$1.44 & 77.84$\pm$0.42 & 67.66$\pm$0.46 & 82.27$\pm$0.89 & \textbf{85.16$\pm$0.73} & \underline{84.60$\pm$0.71}\\
        & $Dir(0.3)$ & 50.97$\pm$3.21 & 70.77$\pm$1.05 & 71.98$\pm$1.87 & 74.35$\pm$1.08 & 66.23$\pm$0.42 & 80.13$\pm$0.72 & \textbf{84.02$\pm$0.46} & \underline{83.11$\pm$0.06}\\
        & $Dir(0.1)$ & 37.91$\pm$3.94 & 55.39$\pm$2.68 & 57.31$\pm$1.37 & 63.22$\pm$0.71 & 64.41$\pm$0.19 & 77.46$\pm$0.54 & \textbf{81.09$\pm$0.56} & \underline{80.53$\pm$0.30}\\

        \cmidrule(lr){1-10}
        \multirow{3}{*}{EMNIST}
        & $Dir(0.5)$ & 61.65$\pm$1.50 & 64.52$\pm$1.40 & 68.13$\pm$0.48 & 68.90$\pm$0.28 & 72.33$\pm$1.23 & 75.73$\pm$0.45 & \textbf{78.45$\pm$0.27} & \underline{77.79$\pm$0.45} \\
        & $Dir(0.3)$ & 57.37$\pm$1.49 & 62.08$\pm$0.52 & 63.54$\pm$0.66 & 66.16$\pm$0.37 & 69.07$\pm$0.72 & 72.63$\pm$0.36 & \underline{74.81$\pm$0.56} & \textbf{75.36$\pm$0.51} \\
        & $Dir(0.1)$ & 39.69$\pm$3.18 & 45.09$\pm$2.52 & 49.07$\pm$1.45 & 58.74$\pm$0.72 & 66.26$\pm$1.21 & 70.85$\pm$0.85 & \underline{73.23$\pm$0.86} & \textbf{73.44$\pm$0.78} \\

        \bottomrule
    \end{tabular}
\end{table*}

\begin{figure*}[t]
    \centering
    
    \subfigure[$Dir(0.5)$]{
    \includegraphics[width=0.31\textwidth]{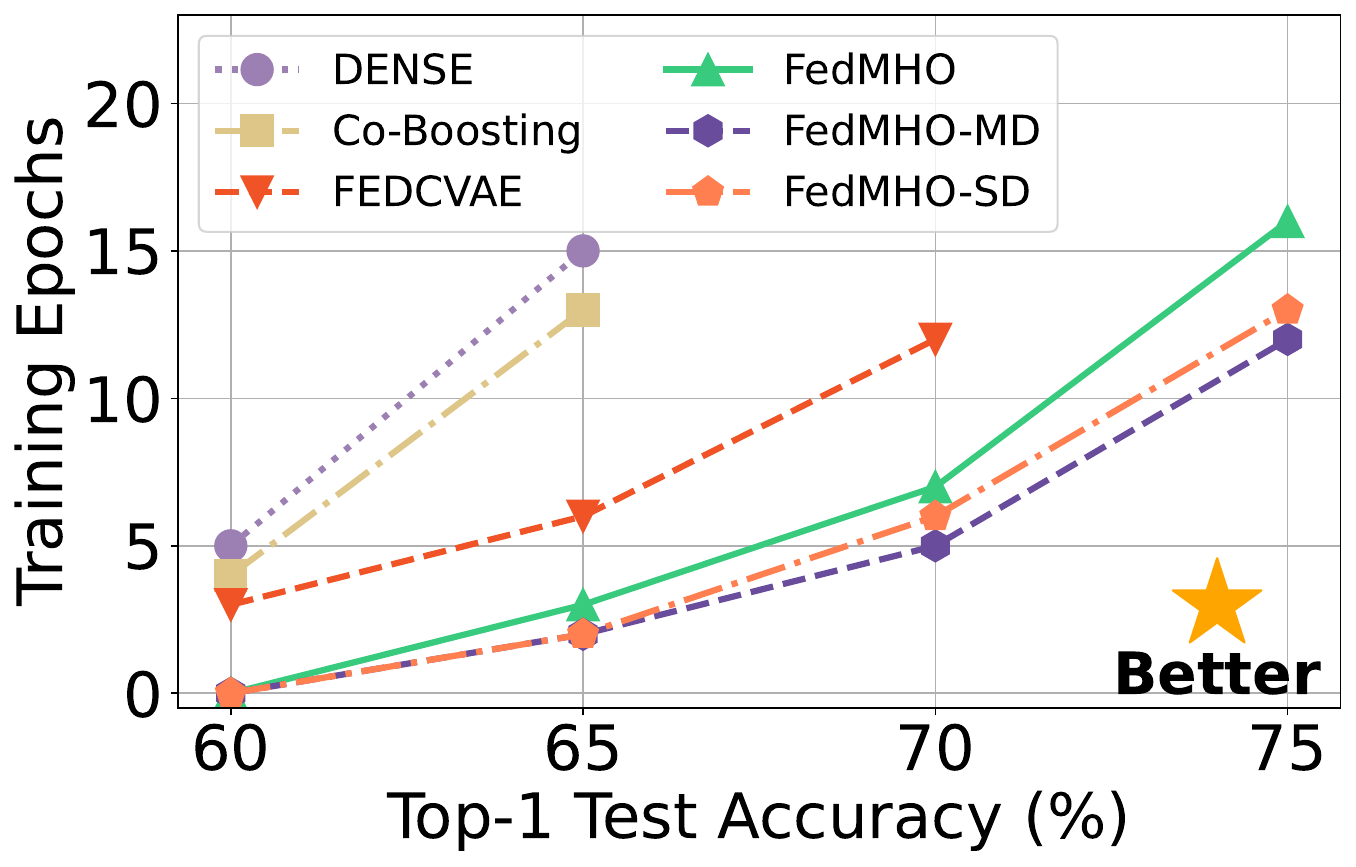}
    }
    \subfigure[$Dir(0.3)$]{
    \includegraphics[width=0.31\textwidth]{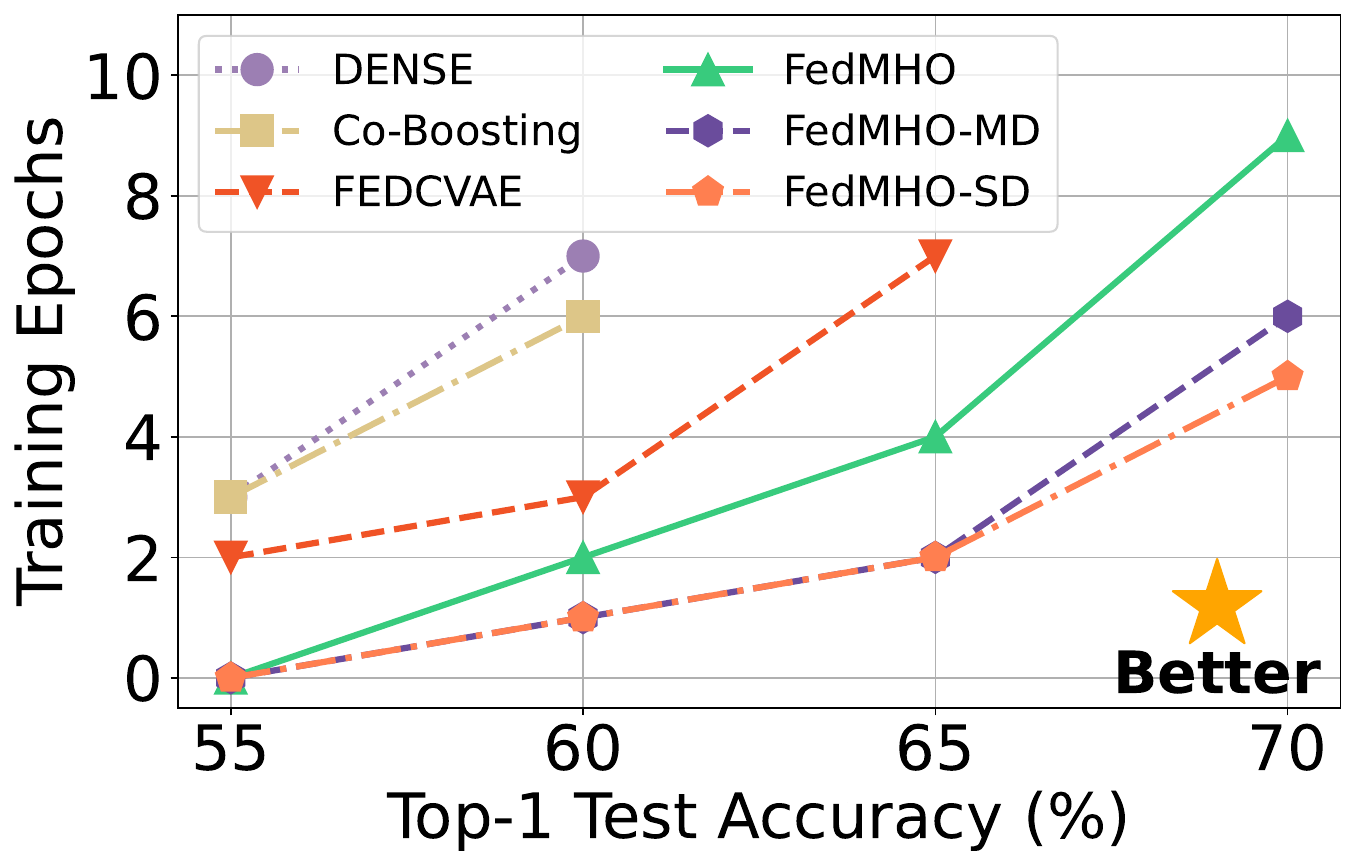}
    }
    \subfigure[$Dir(0.1)$]{
    \includegraphics[width=0.31\textwidth]{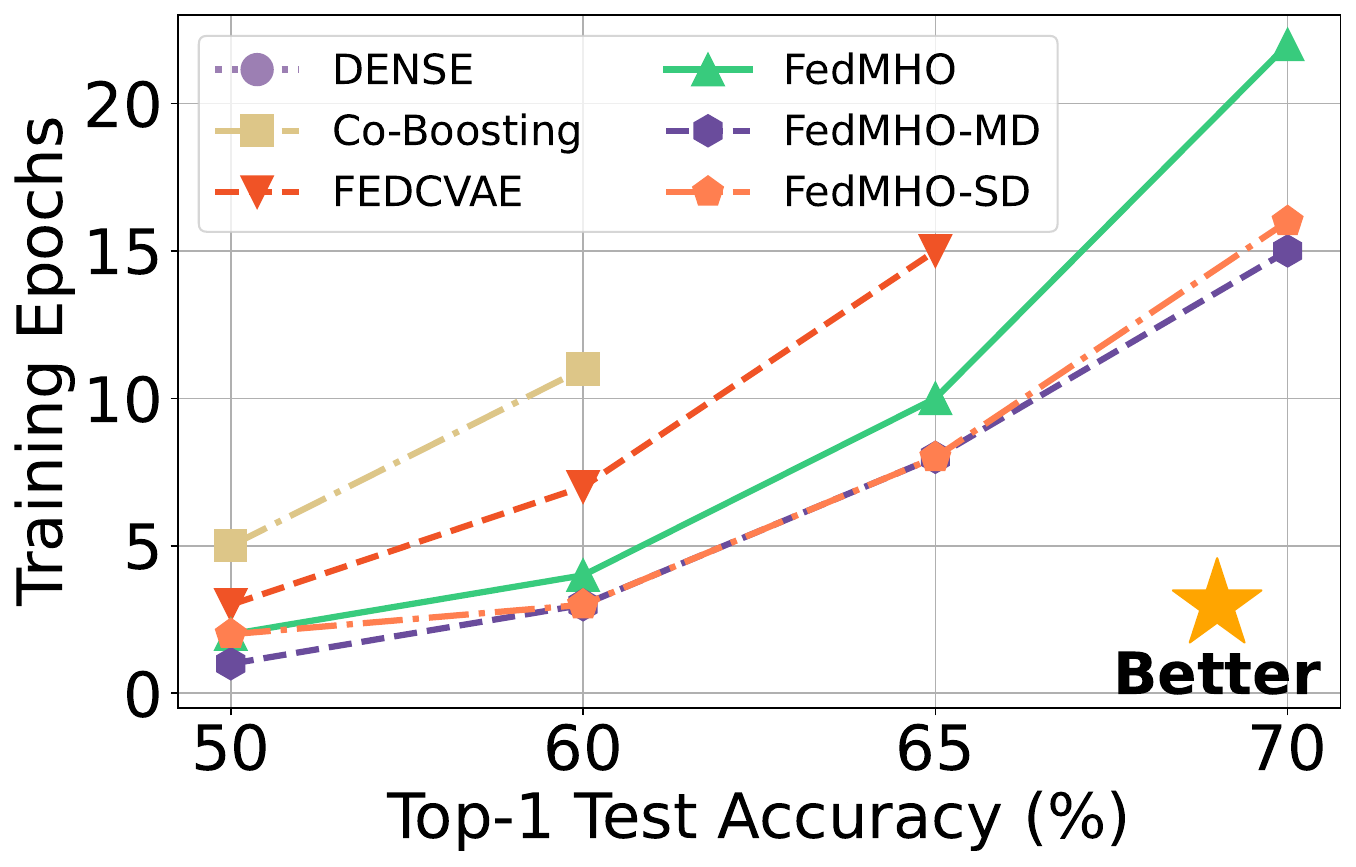}
    }
    \caption{Global training epochs to reach target Top-1 test accuracy on the EMNIST dataset.}
    \label{figure_target_acc}
\end{figure*}

\subsection{Experimental Setup}

\subsubsection{Datasets}
Following the latest baseline, FEDCVAE~\cite{heinbaugh2023datafree}, we evaluate our methods on four widely used datasets: MNIST~\cite{lecun1998gradient}, Fashion-MNIST~\cite{xiao2017fashion} (abbreviated as Fashion), SVHN~\cite{yuval2011reading} and EMNIST~\cite{cohen2017emnist}. 
To simulate statistical heterogeneity, we employ the Dirichlet distribution as in~\cite{hsu2019measuring}, which is denoted as $Dir(\alpha)$. Smaller values of $\alpha$ correspond to greater heterogeneity among clients’ local data partitions. We set $\alpha=\{0.5, 0.3, 0.1\}$ by default.

\subsubsection{Methods}
Under the constraint of a single communication round, FL methods that emphasize regularization, such as FedProx~\cite{li2020federated}, Scaffold~\cite{karimireddy2020scaffold} and FedGen~\cite{zhu2021data}, are rendered ineffective. Given the data-free nature of our proposed methods, we select three state-of-the-art data-free one-shot FL methods as baselines: DENSE~\cite{zhang2022dense}, Co-Boosting~\cite{dai2024enhancing} and FEDCVAE~\cite{heinbaugh2023datafree}. It is noteworthy that FEDCVAE-KD, proposed in~\cite{heinbaugh2023datafree} alongside FEDCVAE-ENS, consistently underperforms compared to FEDCVAE-ENS. Therefore, our comparison focuses exclusively on FEDCVAE-ENS, referred to simply as FEDCVAE. Additionally, we compare our methods with two well-known baselines: FedAvg~\cite{mcmahan2017communication} and FedDF~\cite{lin2020ensemble}. In our experiments, FedMHO represents the scenario without addressing the knowledge-forgetting issue. We further define FedMHO-MD and FedMHO-SD to signify the incorporation of two distinct strategies for mitigating knowledge-forgetting. For simplicity, we refer to the set comprising FedMHO, FedMHO-MD, and FedMHO-SD as FedMHOs. To ensure a fair comparison under the heterogeneous one-shot setting, we limit the communication to a single round for both FedAvg and FedDF, and utilize the voting results of each model prototype as the experimental outcome. A brief introduction to the baseline methods is provided below.
\begin{itemize}
    \item FedAvg~\cite{mcmahan2017communication} is a foundational FL method that learns a global model by averaging the parameters of local models.
    \item FedDF~\cite{lin2020ensemble} employs auxiliary public datasets on the server side to ensemble distill the knowledge of the local models into the global model.
    \item DENSE~\cite{zhang2022dense} leverages local models to train a generator on the server side. After generating synthetic data, it aggregates the local models similarly to FedDF.
    \item Co-Boosting~\cite{dai2024enhancing} is an extension of DENSE. In each round of server-side training, it assigns different weights to the local model based on the feedback of the global model, thereby training more effective generators and improving the global model.
    \item FEDCVAE~\cite{heinbaugh2023datafree} trains generative models on each client and leverages the synthetic data to train the global model.
\end{itemize}


\begin{figure*}[t]
    \centering
    
    \subfigure[$Dir(0.5)$]{
    \includegraphics[width=0.31\textwidth]{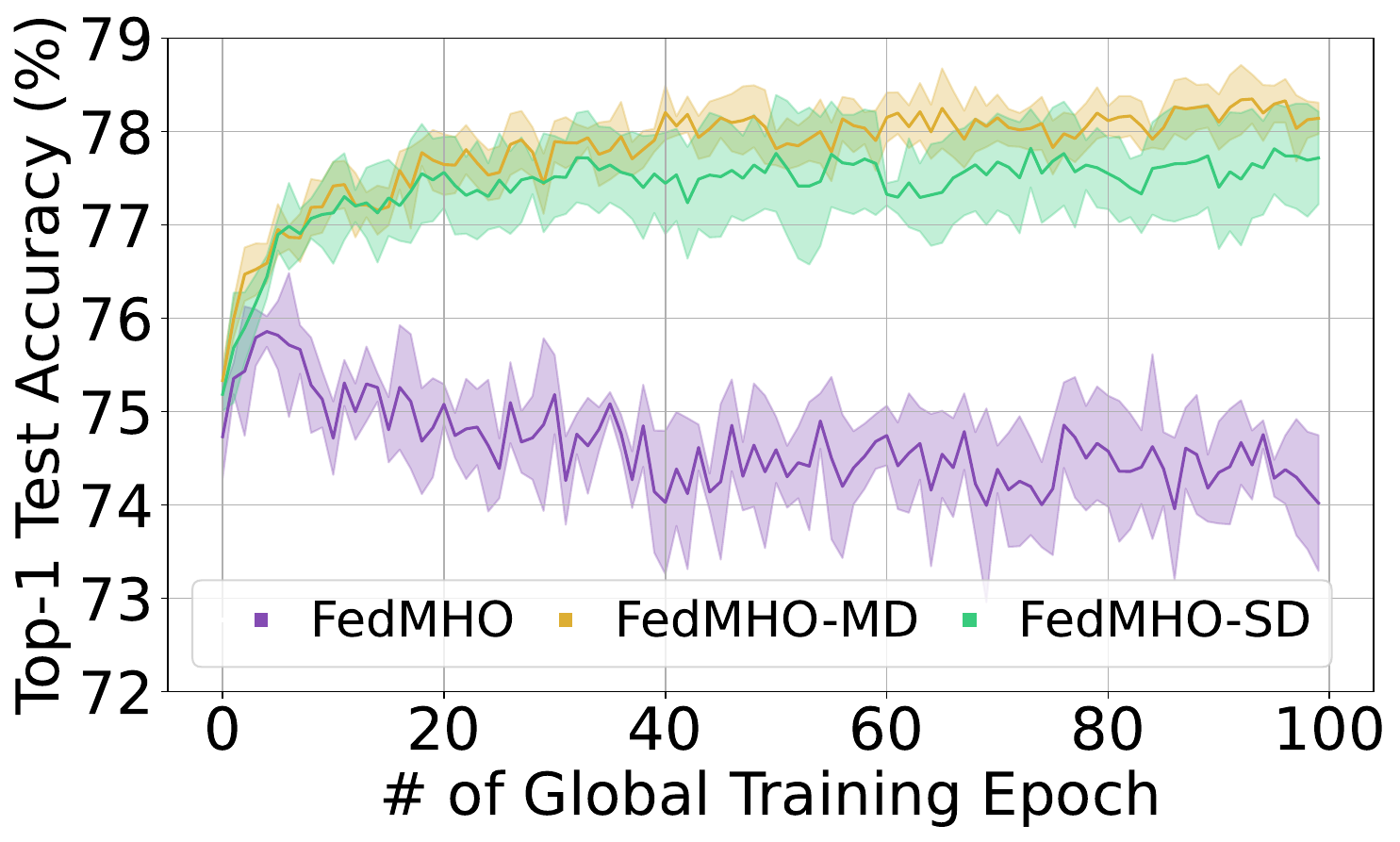}
    }
    \subfigure[$Dir(0.3)$]{
    \includegraphics[width=0.31\textwidth]{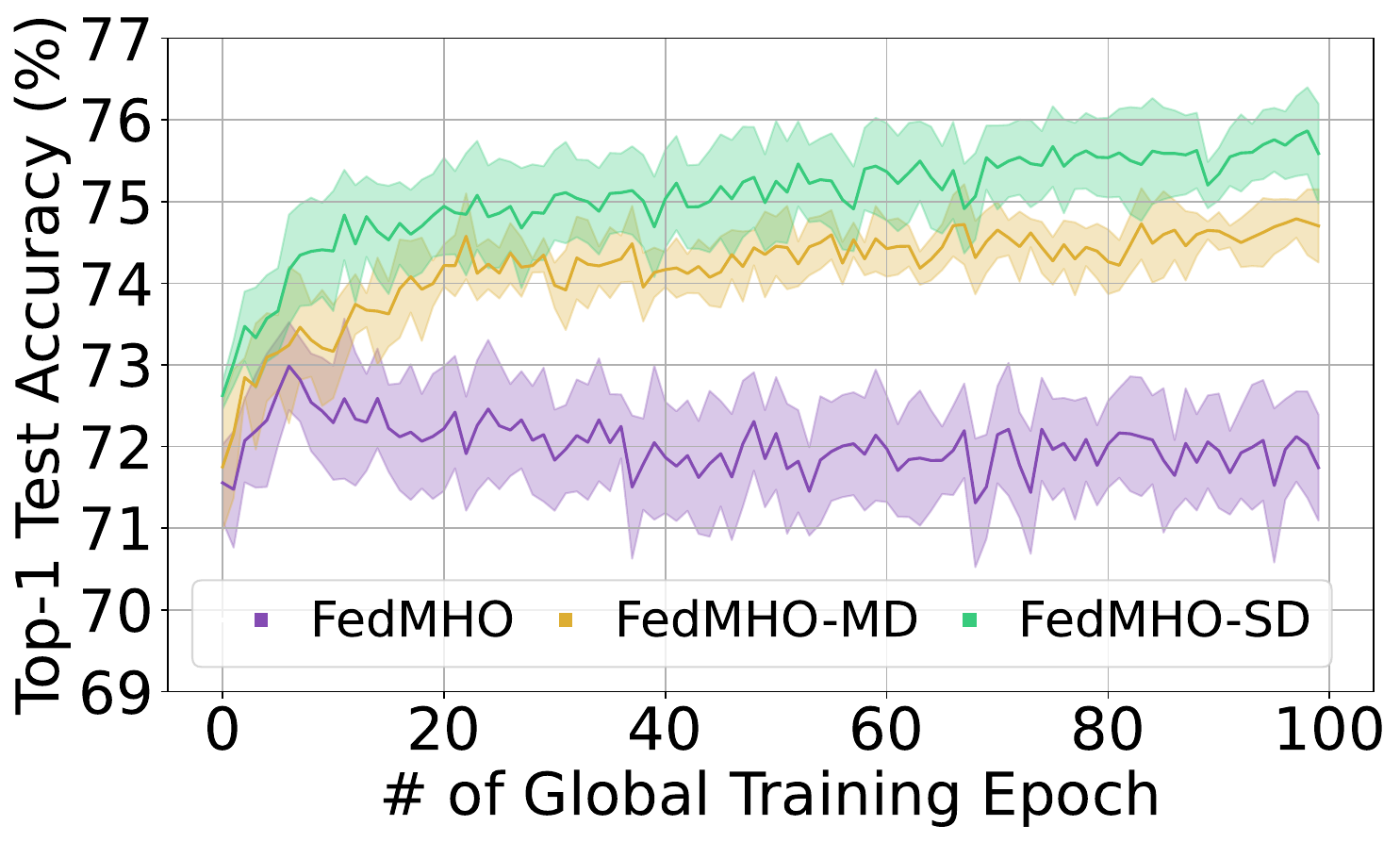}
    }
    \subfigure[$Dir(0.1)$]{
    \includegraphics[width=0.31\textwidth]{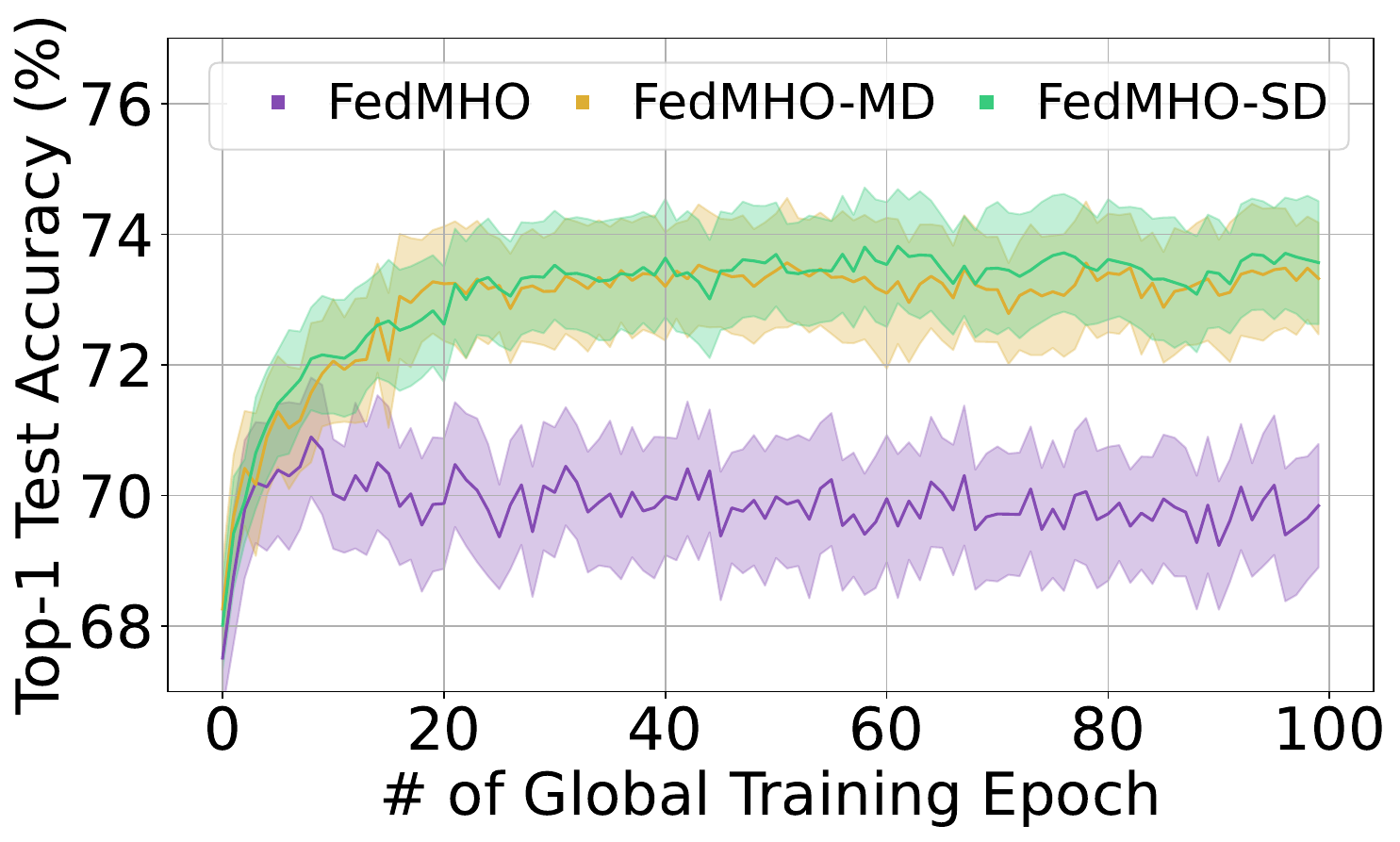}
    }
    \caption{Top-1 test accuracy (\%) curve of FedMHOs on the EMNIST dataset.}
    \label{figure_knowledge_forgetting}
\end{figure*}

\subsubsection{Models}
To simulate model-heterogeneous scenarios, we employ two distinct local model prototypes. For the models with more parameters, we utilize VGG-9 and CVAE-large, while for the models with less parameters, we employ CNN and CVAE-small. This selection is designed to reflect the computing power disparities typical in real-world applications, such as those between smartphones and wristbands.
Table~\ref{table_local_model} provides a summary of the specific local models used by each method, and Table~\ref{table_flops} details the Floating Point Operations (FLOPs) for each local model with a batch size of 1. To demonstrate the efficiency and feasibility of our proposed FedMHOs in resource-constrained environments, our methods consistently train the local models that have the lowest FLOPs. In our evaluation, we report the global model's Top-1 test accuracy in the form of \verb+mean+ $\pm$ \verb+standard deviation+, ensuring the reliability of the results.

\subsubsection{Configurations}
For training local classification models, we utilize the SGD optimizer with a momentum of 0.9 and a learning rate of 5$e$-3, training for 200 local epochs. For local generative models, we use the Adam optimizer with varying learning rates specific to each dataset: 5$e$-2 for MNIST and Fashion, 1$e$-3 for SVHN, and 3$e$-3 for EMNIST, with training conducted over 30, 40, 40, and 50 local epochs respectively. When training the global model, we use the Adam optimizer with a learning rate of 1$e$-5, 5$e$-4, 5$e$-5, and 5$e$-5 for MNIST, Fashion, SVHN, and EMNIST respectively. For methods involving generators, we produce 6,000 synthetic samples each for MNIST, Fashion, and SVHN, and 12,000 synthetic samples for EMNIST. Consistent with the latest baseline FEDCVAE~\cite{heinbaugh2023datafree}, we configure the FL setup with 10 clients participating in the training process. By default, 5 clients simulate resource-sufficient clients, and the remaining 5 simulate resource-constrained clients. The detailed hyperparameter settings are provided in Table~\ref{table_hyperparameter}.

\begin{table}[t]
    \normalsize
    \caption{Top-1 test accuracy (\%) of different weights of loss function on EMNIST dataset. $\mathcal{l}$ indicates learnable.}
    \label{table_loss_weight}
    \addtolength{\tabcolsep}{-2pt}
    \centering
    \begin{tabular}{c|c|c|cc}
    \toprule
        $\beta_1$ & $\beta_2$ & Partition & FedMHO-MD & FedMHO-SD \\

        \cmidrule(lr){1-5}
        \multirow{3}{*}{1} & 
        & $Dir(0.5)$ & 78.45$\pm$0.27 & 77.79$\pm$0.45 \\
        & 1 & $Dir(0.3)$ & 74.81$\pm$0.56 & 75.36$\pm$0.51 \\
        & & $Dir(0.1)$ & 73.23$\pm$0.86 & 73.44$\pm$0.78 \\

        \cmidrule(lr){1-5}
        \multirow{3}{*}{1} & 
        & $Dir(0.5)$ & 77.86$\pm$0.35 (-0.59) & 77.34$\pm$0.47 (-0.45) \\
        & 2 & $Dir(0.3)$ & 74.44$\pm$0.42 (-0.37) & 75.03$\pm$0.63 (-0.33) \\
        & & $Dir(0.1)$ & 72.80$\pm$0.69 (-0.43) & 73.15$\pm$0.39 (-0.29) \\

        \cmidrule(lr){1-5}
        \multirow{3}{*}{2} & 
        & $Dir(0.5)$ & 78.18$\pm$0.14 (-0.27) & 77.65$\pm$0.56 (-0.14) \\
        & 1 & $Dir(0.3)$ & 74.69$\pm$0.45 (-0.12) & 75.22$\pm$0.66 (-0.14) \\
        & & $Dir(0.1)$ & 73.24$\pm$0.48 (+0.01) & 73.38$\pm$0.73 (-0.06) \\

        \cmidrule(lr){1-5}
        \multirow{3}{*}{$\mathcal{l}$} & 
        & $Dir(0.5)$ & 78.36$\pm$0.33 (-0.09) & 77.85$\pm$0.40 (-0.06) \\
        & $\mathcal{l}$ & $Dir(0.3)$ & 75.06$\pm$0.60 (+0.25) & 75.18$\pm$0.64 (-0.18) \\
        & & $Dir(0.1)$ & 73.18$\pm$0.42 (-0.05) & 73.21$\pm$0.43 (-0.23) \\
        
        \bottomrule
    \end{tabular}
\end{table}

\subsection{Performance Comparison}
\label{section_performance_comparison}
We present a comprehensive comparison of our proposed FedMHOs with the baselines in Table~\ref{table_performance_comparison}. The results demonstrate that our methods outperform the baselines across various datasets and data partitions, with FedMHO-MD and FedMHO-SD consistently achieving the top two rankings in Top-1 test accuracy. Specifically, FedMHO-MD and FedMHO-SD achieve the highest Top-1 test accuracy in 6 out of 12 experiments, respectively. Under the $Dir(0.5)$ local data partition, FedMHO-MD or FedMHO-SD achieves superior results compared to the best-performing baseline. The improvements are 3.06\%, 5.44\%, 7.23\%, and 6.12\% on the MNIST, Fashion, SVHN, and EMNIST datasets, respectively.

Even in the absence of the loss function $\mathcal{L}_{\text{KD}}$, which mitigates the knowledge-forgetting problem of the global model, FedMHO generally outperforms other baselines, except when training on the MNIST and Fashion datasets with the $Dir(0.1)$ local data partition. We attribute this to the pronounced knowledge deviation between generative and classification local models trained on highly heterogeneous local data. This deviation intensifies the effect of knowledge-forgetting when the global model is trained on synthetic samples. While the best-performing baseline, FEDCVAE, outperforms FedMHO under highly non-IID local data partitions, it fails to surpass FedMHO-MD or FedMHO-SD. 
Additionally, since FedMHOs' global models are well initialized, FedMHOs converge faster than the baselines during global model training, as shown in Figure~\ref{figure_target_acc}. Overall, our experimental results demonstrate the effectiveness and robustness of our proposed FedMHOs, highlighting their potential for real-world applications.

\begin{table}[t]
    \normalsize
    \caption{Top-1 test accuracy (\%) of FedMHO-MD, FedMHO-SD without utilizing unsupervised data optimization. }
    \label{table_without_unsupervised_data_optimization}
    \centering
    \begin{tabular}{c|c|cc}
    \toprule
        Dataset & Partition & FedMHO-MD & FedMHO-SD \\

        \cmidrule(lr){1-4}
        \multirow{3}{*}{MNIST}
        & $Dir(0.5)$ & 94.45$\pm$0.42 & 94.70$\pm$0.38 \\
        & $Dir(0.3)$ & 92.33$\pm$0.23 & 93.17$\pm$0.32 \\
        & $Dir(0.1)$ & 90.66$\pm$0.31 & 90.33$\pm$0.86\\

        \cmidrule(lr){1-4}
        \multirow{3}{*}{Fashion}
        & $Dir(0.5)$ & 74.56$\pm$1.02 & 75.11$\pm$1.93 \\
        & $Dir(0.3)$ & 73.15$\pm$0.52 & 74.12$\pm$2.40 \\
        & $Dir(0.1)$ & 65.13$\pm$0.91 & 65.42$\pm$3.68 \\

        \cmidrule(lr){1-4}
        \multirow{3}{*}{SVHN}
        & $Dir(0.5)$ & 84.63$\pm$0.72 & 84.11$\pm$0.93\\
        & $Dir(0.3)$ & 83.23$\pm$1.09 & 82.54$\pm$0.32\\
        & $Dir(0.1)$ & 80.27$\pm$0.86 & 79.82$\pm$1.15\\

        \cmidrule(lr){1-4}
        \multirow{3}{*}{EMNIST}
        & $Dir(0.5)$ & 76.79$\pm$0.18 & 76.86$\pm$0.33 \\
        & $Dir(0.3)$ & 74.33$\pm$0.68 & 74.89$\pm$0.47 \\
        & $Dir(0.1)$ & 72.81$\pm$0.40 & 72.84$\pm$0.51 \\
        
        \bottomrule
    \end{tabular}
\end{table}

\begin{figure*}[t]
    \centering
    
    \subfigure[$Dir(0.5)$]{
    \includegraphics[width=0.31\textwidth]{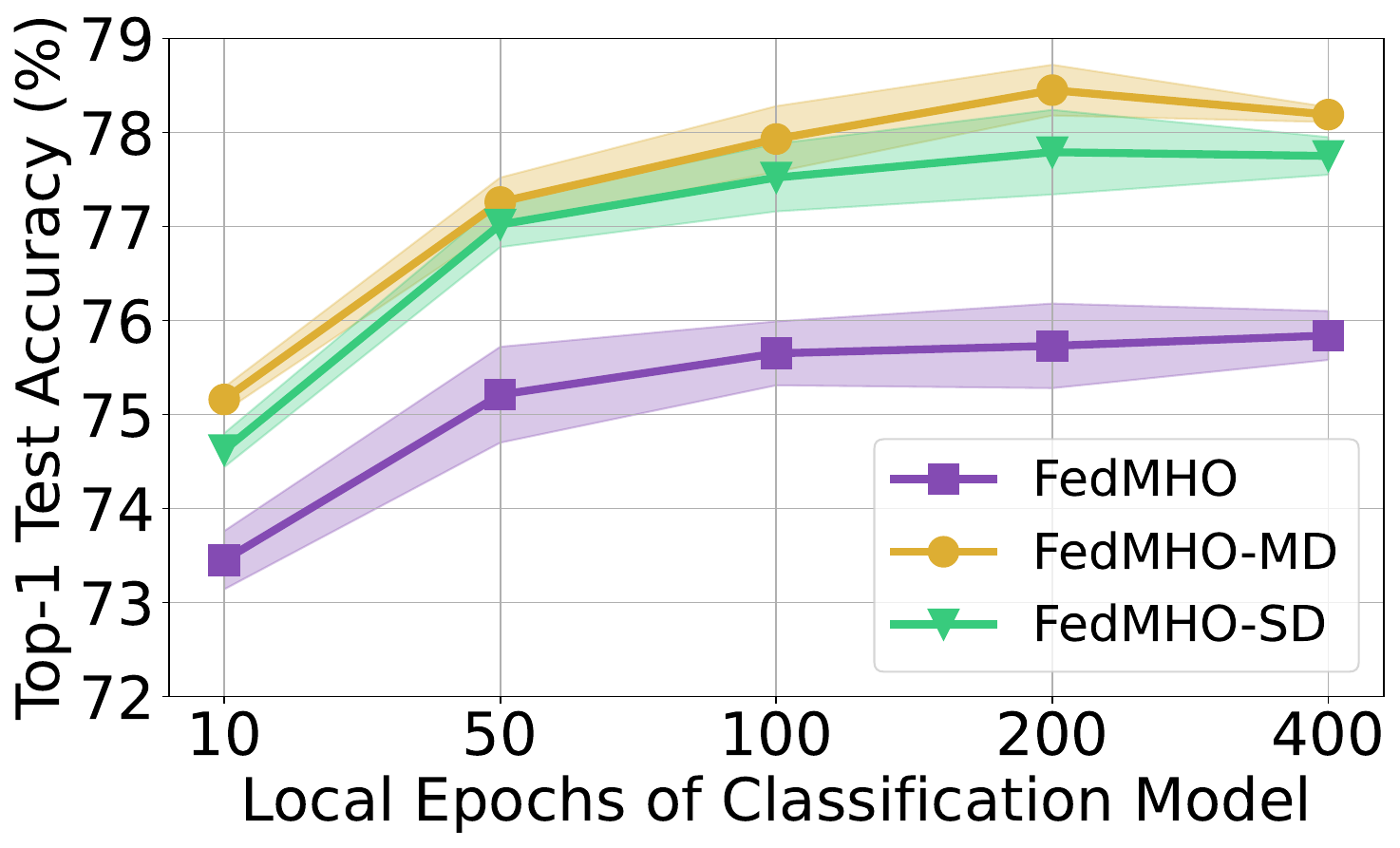}
    }
    \subfigure[$Dir(0.3)$]{
    \includegraphics[width=0.31\textwidth]{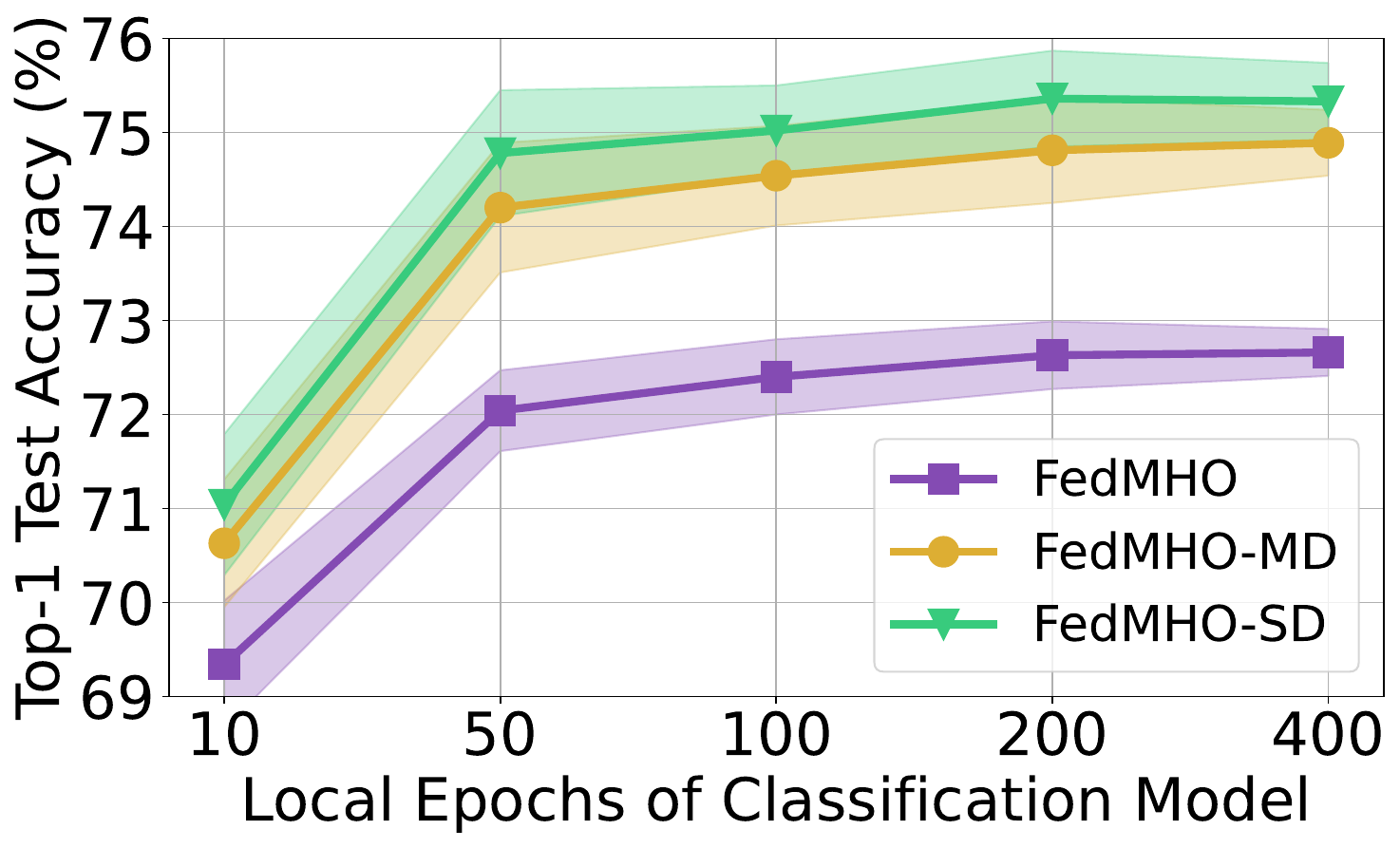}
    }
    \subfigure[$Dir(0.1)$]{
    \includegraphics[width=0.31\textwidth]{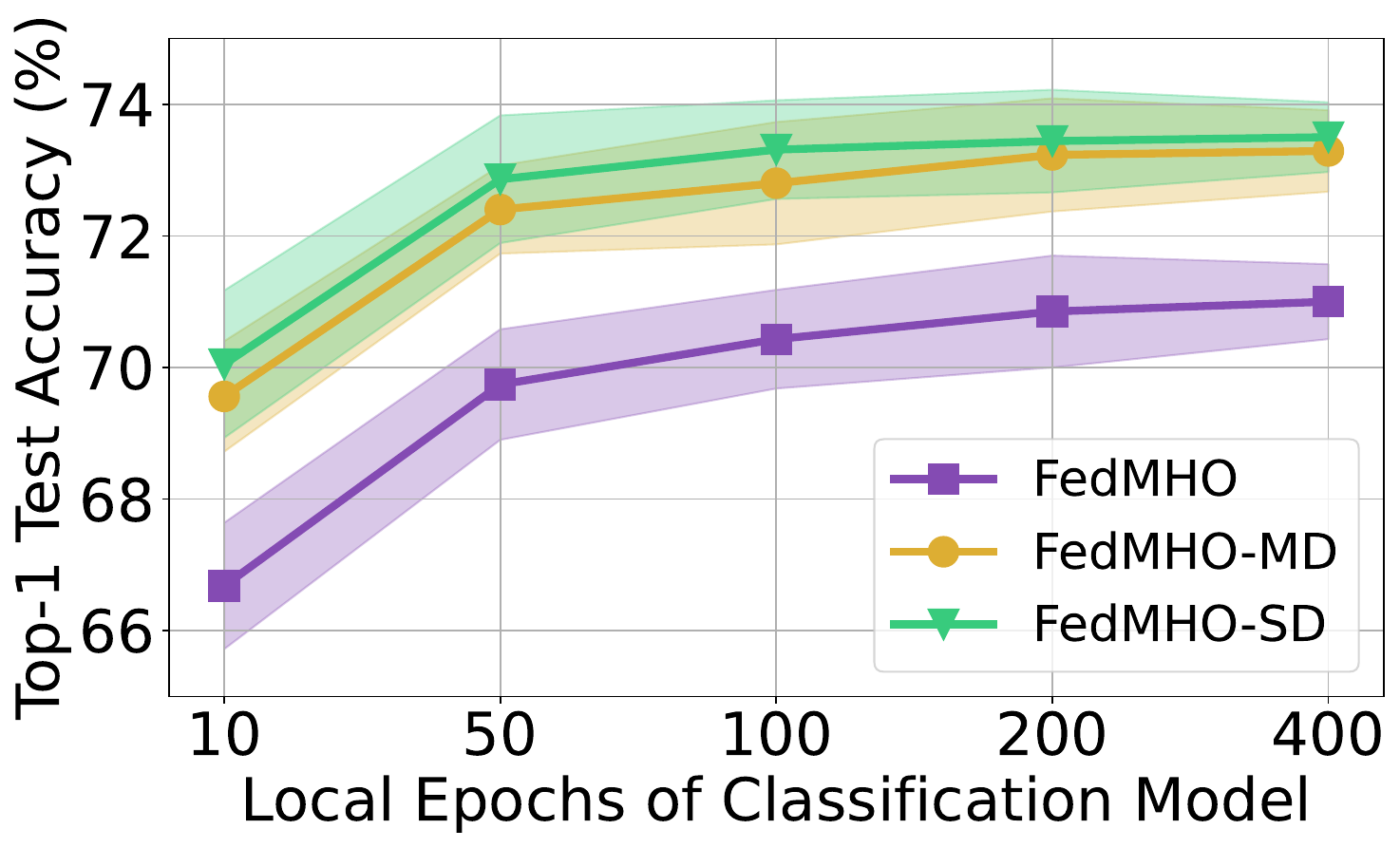}
    }
    
    \caption{Top-1 test accuracy (\%) of FedMHOs on the EMNIST dataset under various numbers of local epoch of classification models.}
    \label{figure_number_local_epoch}
\end{figure*}

\begin{figure*}[t]
    \centering
    
    \subfigure[FedMHO-MD, $Dir(0.5)$]{
    \includegraphics[width=0.31\textwidth]{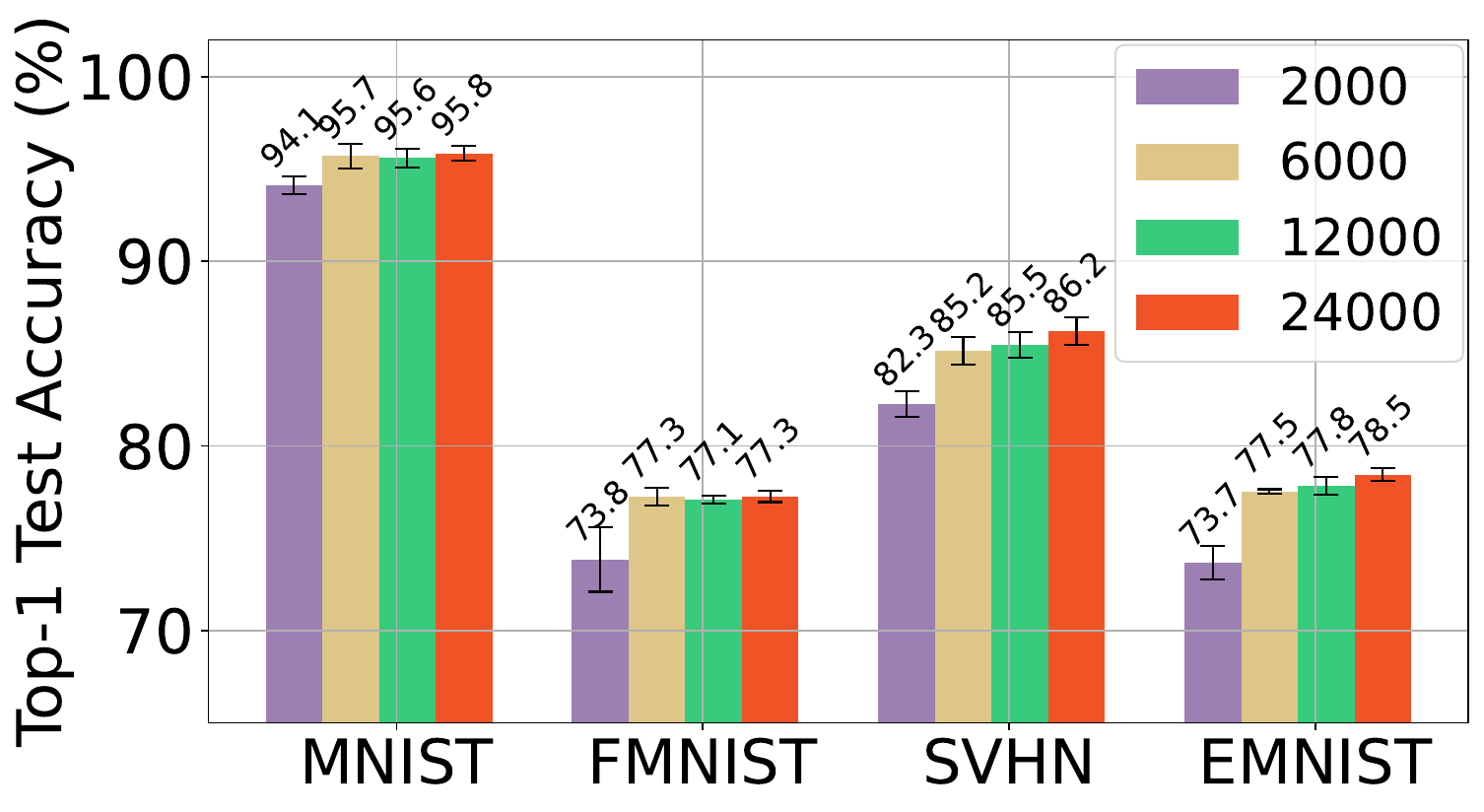}
    }
    \subfigure[FedMHO-MD, $Dir(0.3)$]{
    \includegraphics[width=0.31\textwidth]{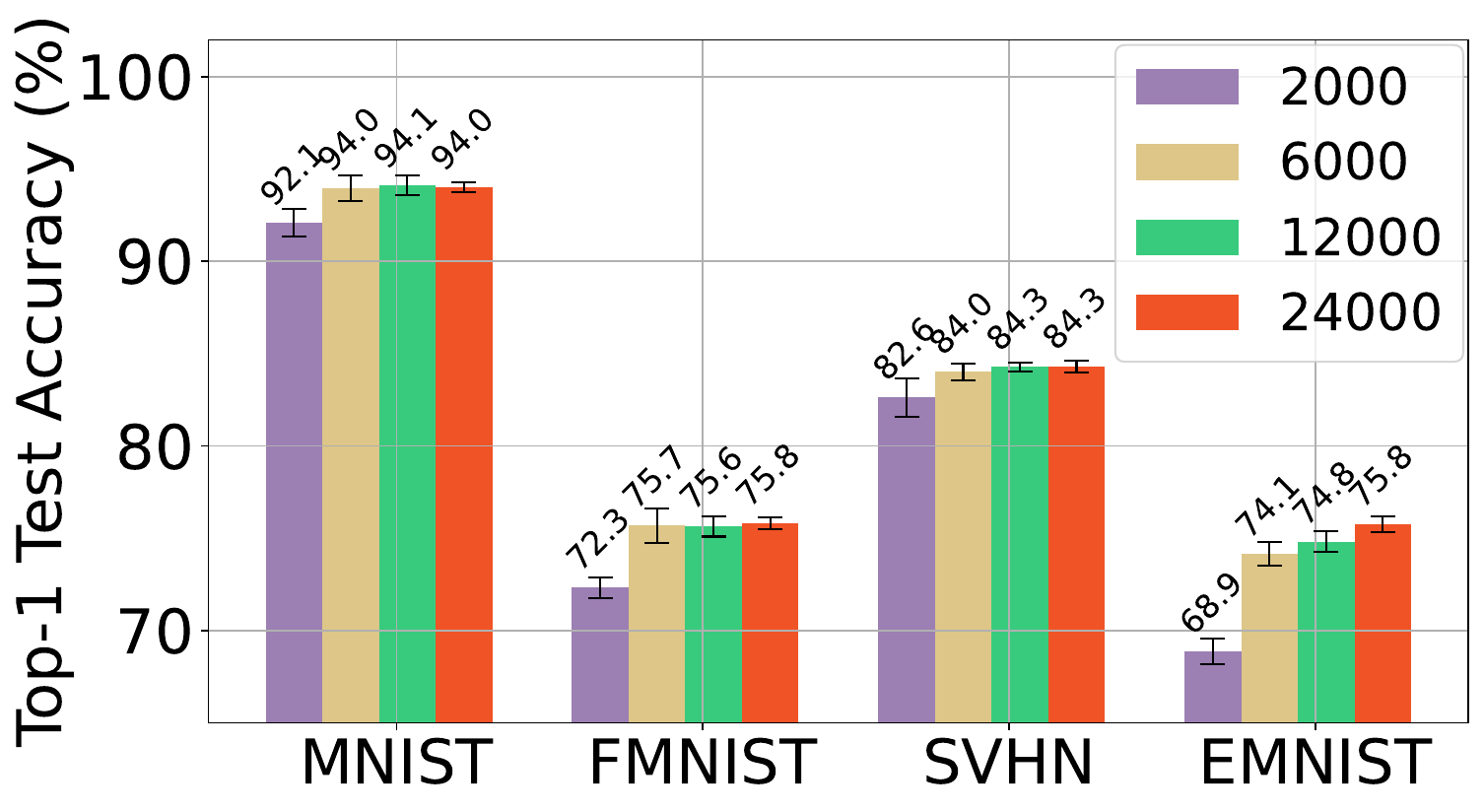}
    
    }
    \subfigure[FedMHO-MD, $Dir(0.1)$]{
    \includegraphics[width=0.31\textwidth]{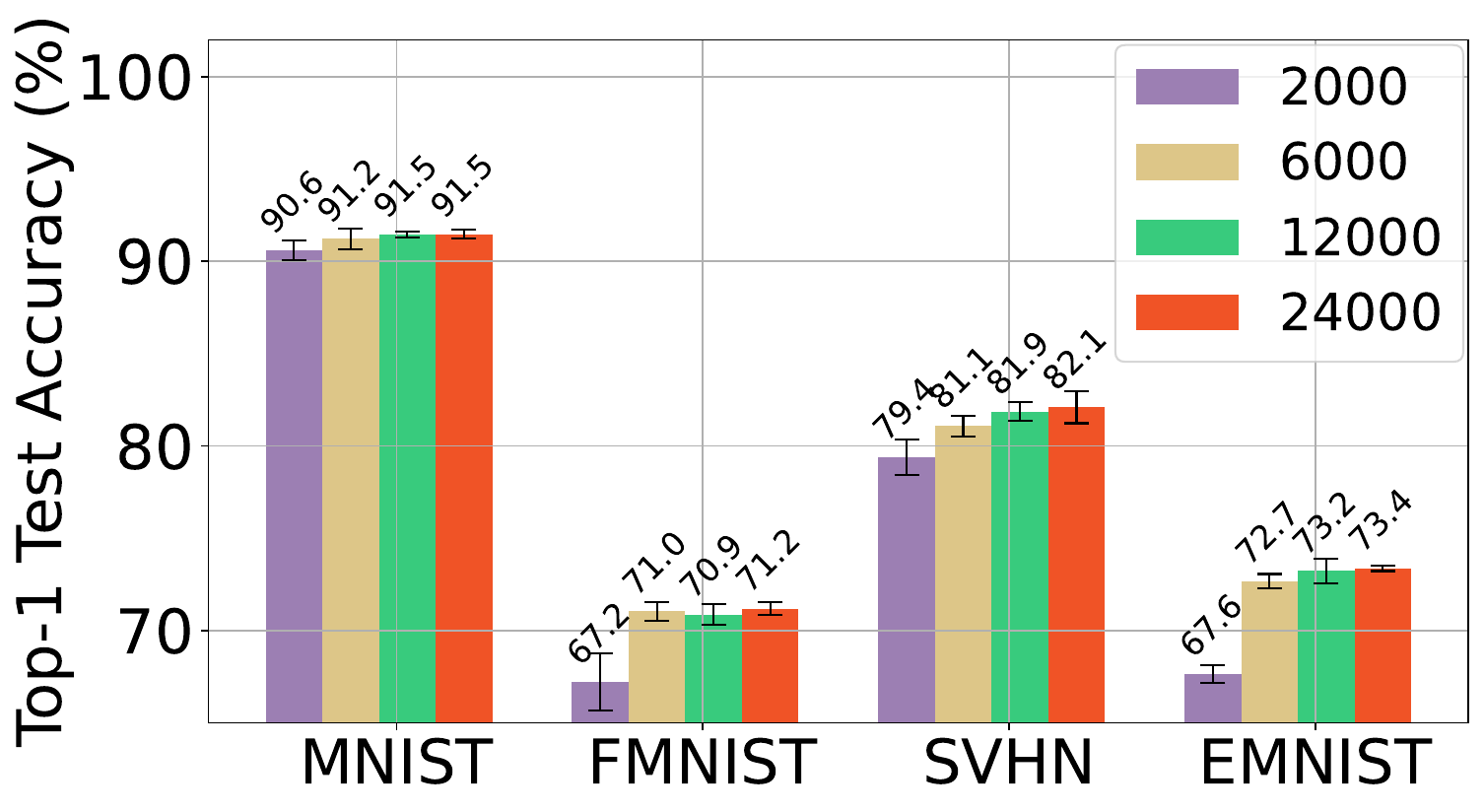}
    }

    \subfigure[FedMHO-SD, $Dir(0.5)$]{
    \includegraphics[width=0.31\textwidth]{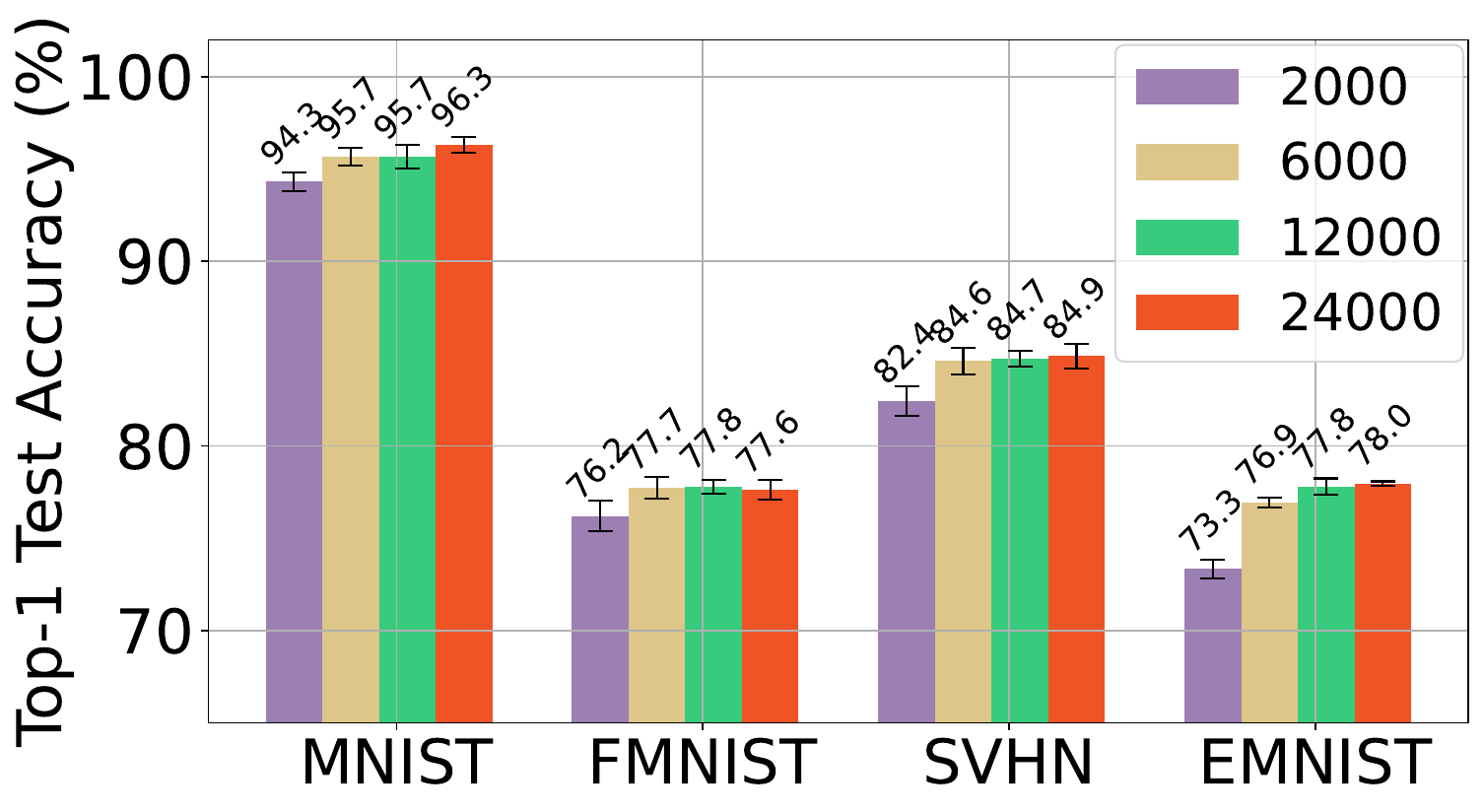}
    }
    \subfigure[FedMHO-SD, $Dir(0.3)$]{
    \includegraphics[width=0.31\textwidth]{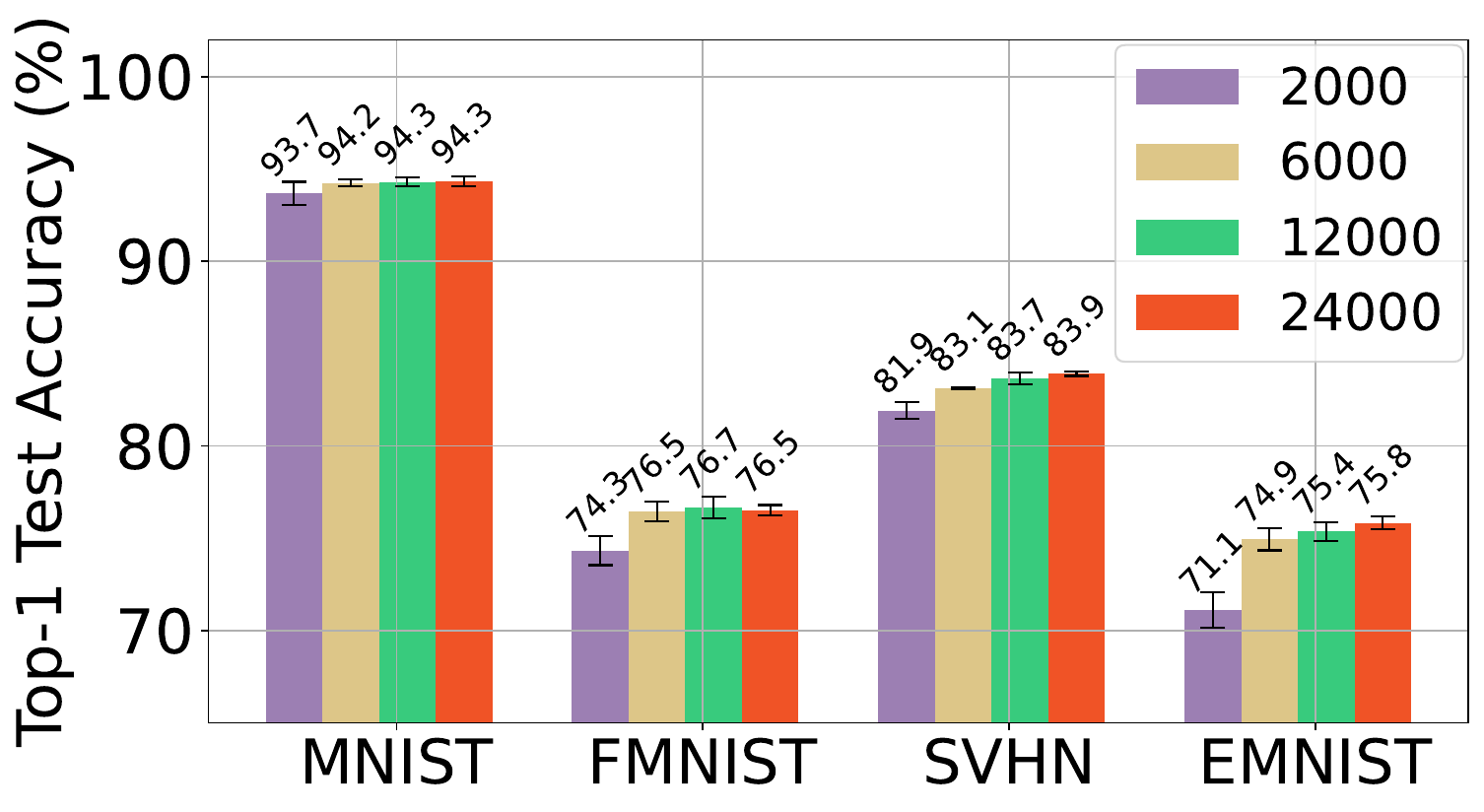}
    }
    \subfigure[FedMHO-SD, $Dir(0.1)$]{
    \includegraphics[width=0.31\textwidth]{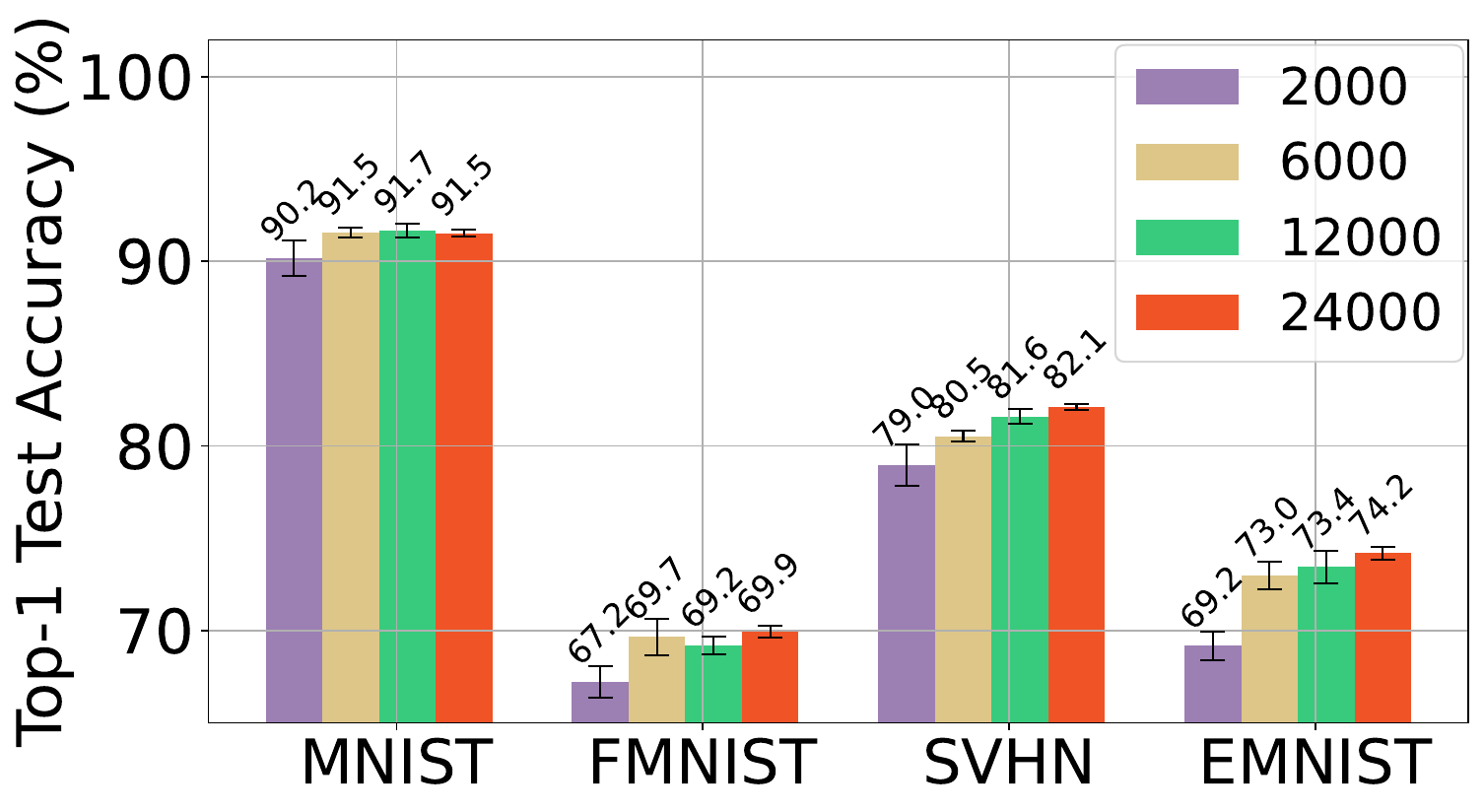}
    }
    \caption{Top-1 test accuracy (\%) of FedMHO-MD and FedMHO-SD under various number of synthetic samples.}
    \label{figure_number_synthetic_samples}
\end{figure*}

\subsection{Ablation Study}
\subsubsection{Contribution of $\mathcal{L}_{\text{KD}}$}
As mentioned in Section~\ref{section_knowledge_fusion}, we employ the loss function $\mathcal{L}_{\text{KD}}$ to mitigate knowledge-forgetting during global model training. Table~\ref{table_performance_comparison} illustrates that both FedMHO-MD and FedMHO-SD benefit from $\mathcal{L}_{\text{KD}}$, improving the global model's Top-1 test accuracy. To further substantiate the necessity of $\mathcal{L}_{\text{KD}}$, we extend the training epochs of the global model to 100 epochs. The variation of Top-1 test accuracy throughout the training process is depicted in Figure~\ref{figure_knowledge_forgetting}. As shown in the figure, in the absence of $\mathcal{L}_{\text{KD}}$, the global model acquires new knowledge from the synthetic data but forgets the original knowledge obtained from aggregating the classification models. This finding underscores the importance of addressing knowledge-forgetting.

To further analyze the experimental results under different weight configurations, we define the weight of $\mathcal{L}_\text{KD}$ as $\beta_1$ and the weight of $\mathcal{L}_\text{CE}$ as $\beta_2$. By default, both $\beta_1$ and $\beta_2$ are set to 1. Table~\ref{table_loss_weight} summarizes the experimental results on the EMNIST dataset under various $\beta_1$ and $\beta_2$. When $\beta_1=1,\beta_2=2$ or $\beta_1=2,\beta_2=1$, the global model's Top-1 test accuracy decreases slightly. Moreover, when $\beta_1$ and $\beta_2$ are treated as trainable parameters, the performance remains comparable to the default setting of $\beta_1=1$ and $\beta_2=1$. Given that treating the weights as trainable parameters introduces additional computation overhead and potential optimization challenges, we choose the default configuration as the final solution.

Additionally, we investigate the impact of simultaneously utilizing the loss functions associated with FedMHO-MD's multi-teacher knowledge distillation and FedMHO-SD's self-distillation. We conduct experiments on the EMNIST dataset that satisfies the $Dir(0.5)$ data partition. When the weights of both two $\mathcal{L}_\text{KD}$s are set to 1, the Top-1 test accuracy is 78.14\%. When both weights are treated as trainable parameters, the Top-1 test accuracy is 78.33\%. These results indicate that the simultaneous use of both $\mathcal{L}\text{KD}$ terms does not lead to substantial performance gains compared to using a single $\mathcal{L}\text{KD}$. This outcome can be attributed to the fact that both $\mathcal{L}\text{KD}$s aim to mitigate knowledge forgetting by regularizing the global model based on the local classification models. Thus, a single $\mathcal{L}_\text{KD}$ suffices to achieve this objective.

\subsubsection{Contribution of Unsupervised Data Optimization}
\label{section_contribution_of_unsupervised_data_optimization}
To demonstrate the impact of unsupervised data optimization, we present the experimental results obtained without this technique in Table~\ref{table_without_unsupervised_data_optimization}. By default, our methods retain 80\% of generated data when utilizing unsupervised data optimization, resulting in the use of 4,800 synthetic samples for MNIST, Fashion, and SVHN, and 96,000 for EMNIST in global model training. To ensure a fair comparison, we use an equivalent number of synthetic samples without unsupervised data optimization. Comparison of Table~\ref{table_without_unsupervised_data_optimization} and Table~\ref{table_performance_comparison} shows that the integration of unsupervised data optimization consistently improves the global model performance. For example, under a $Dir(0.1)$ local data partition, FedMHO-MD's Top-1 test accuracy improves by 0.56\%, 5.91\%, 0.82\%, and 0.42\%, while FedMHO-SD’s improves by 1.22\%, 4.23\%, 0.71\%, and 0.60\% on MNIST, Fashion, SVHN, and EMNIST, respectively.

\subsection{Sensitivity analysis}

\subsubsection{Number of Local Epochs}
To verify the impact of the number of local training epochs for the classification model on the experimental results, we vary the number of local epochs for VGG-9 among $\{$10, 50, 200, 400$\}$ and plot the resulting Top-1 test accuracy of the global model in Figure~\ref{figure_number_local_epoch}. When the number of local training epochs is less than 50 rounds, increasing the number of epochs improves the performance of the local classification model, which provides a strong initialization for the global model and enhancing its final accuracy. When the local epochs exceed 50 rounds, the performance of the global model stabilizes, indicating that the local VGG-9 models have converged. Overall, FedMHOs are insensitive to the number of local classification training rounds and training for 200 rounds by default is reasonable for local classification models.

\subsubsection{Number of Synthetic Samples}
To study the effect of the number of synthetic samples, we vary the number of synthetic samples among $\{$2,000, 6,000, 12,000, 24,000$\}$ and present the experimental results in Figure~\ref{figure_number_synthetic_samples}. We observe that as the number of synthetic samples increases, the Top-1 test accuracy of the global model improves accordingly. However, the degree of improvement becomes less pronounced once the number of synthetic samples surpasses 6,000. Taking the EMNIST dataset as an example, when the client data distribution satisfies $Dir(0.5)$, the accuracy of FedMHO-MD and FedMHO-SD increases by 3.8\% and 2.6\%, respectively when the number of synthetic samples increases from 2,000 to 6,000. However, when the number of synthetic samples increases from 6,000 to 24,000, the accuracy of FedMHO-MD and FedMHO-SD only increases by 1.0\% and 1.1\%, respectively.

\subsubsection{Number of Clients}
To study the robustness of FedMHOs to the number of clients, we vary the number of clients $K$ from the default 10 to 6, 12, and 20, while maintaining an equal number of clients deploying the large model and the small model, such that each group consists of $\frac{K}{2}$ clients. We conduct experiments on the EMNIST dataset and present the results in Table~\ref{table_number_of_clients}. Compared with the default setting of 10 clients, all methods show different degrees of improvement in global model accuracy when the number of clients is reduced to 6. In contrast, as the number of clients increases, especially when it reaches 20, the accuracy of the global model generally decreases. Compared to the state-of-the-art baselines DENSE, Co-Boosting, and FEDCVAE, our proposed FedMHOs maintain their robustness with varying numbers of participating clients.




\begin{table}[t]
    \normalsize
    \caption{Top-1 test accuracy (\%) when training EMNIST on $Dir(0.5)$ data partition with various numbers of clients $K$.}
    \label{table_number_of_clients}
    \centering
    \begin{tabular}{c|ccc}
    \toprule
         Method & $K=6$ & $K=12$ & $K=20$\\

        \cmidrule(lr){1-4}
        DENSE & 72.96$\pm$0.25 & 67.88$\pm$0.45 & 65.73$\pm$0.69\\
        
        Co-Boosting & 73.50$\pm$0.38 & 68.05$\pm$0.24 & 65.46$\pm$0.48\\

        FEDCVAE & 75.29$\pm$0.54 & 72.65$\pm$0.52 & 71.24$\pm$0.83\\

        FedMHO & 76.03$\pm$0.28 & 75.23$\pm$0.52 & 74.01$\pm$0.65\\

        FedMHO-MD & \textbf{78.69$\pm$0.15} & \textbf{78.16$\pm$0.30} & \underline{76.55$\pm$0.56}\\

        FedMHO-SD & \underline{78.12$\pm$0.21} & \underline{77.45$\pm$0.14} & \textbf{75.94$\pm$0.38}\\

        \bottomrule
    \end{tabular}
\end{table}

\begin{table}[t]
    \normalsize
    \caption{Top-1 test accuracy (\%) of FedMHO, FedMHO-MD, FedMHO-SD on EMNIST with various $\mathcal{R}_{th}$. }
    \label{table_ratio_of_data_optimization}
    \centering
    \begin{tabular}{c|c|cc}
    \toprule
        $\mathcal{R}_{th}$ & Partition & FedMHO-MD & FedMHO-SD \\

        \cmidrule(lr){1-4}
        \multirow{3}{*}{$40\%$}
        & $Dir(0.5)$ & 77.58$\pm$0.55 & 77.29$\pm$0.62 \\
        & $Dir(0.3)$ & 74.37$\pm$0.83 & 75.06$\pm$0.52 \\
        & $Dir(0.1)$ & 72.99$\pm$0.95 & 73.06$\pm$0.61\\

        \cmidrule(lr){1-4}
        \multirow{3}{*}{$60\%$}
        & $Dir(0.5)$ & 78.04$\pm$0.44 & 77.41$\pm$0.30 \\
        & $Dir(0.3)$ & 74.60$\pm$0.64 & 75.14$\pm$0.56 \\
        & $Dir(0.1)$ & 73.20$\pm$0.67 & 73.11$\pm$0.54 \\

        \cmidrule(lr){1-4}
        \multirow{3}{*}{$90\%$}
        & $Dir(0.5)$ & 78.12$\pm$0.34 & 77.75$\pm$0.26 \\
        & $Dir(0.3)$ & 74.78$\pm$0.47 & 75.12$\pm$0.38 \\
        & $Dir(0.1)$ & 73.15$\pm$0.31 & 73.29$\pm$0.62 \\
        
        \bottomrule
    \end{tabular}
\end{table}

\subsubsection{Ratio of Data Optimization}
\label{section_ratio_of_data_optimization}
As mentioned in Section~\ref{section_contribution_of_unsupervised_data_optimization}, we propose a strategy for unsupervised data optimization. By default, the retention ratio of the remaining samples relative to the original samples, denoted as $\mathcal{R}_{th}$, is set to 80\%. To investigate the impact of $\mathcal{R}_{th}$ on the experimental results, we perform a series of experiments using the EMNIST dataset. We fix the number of synthetic data at 12,000 and vary the value of $\mathcal{R}_{th}$ among 40\%, 60\%, and 90\%. The experimental results are presented in Table~\ref{table_ratio_of_data_optimization}. Comparing the results with Table~\ref{table_performance_comparison} (where $\mathcal{R}_{th}$ is set to 80\%) and Table~\ref{table_without_unsupervised_data_optimization} (which does not include unsupervised data optimization), we find that our default setting of 80\% is optimal. Retaining too high a proportion ($\mathcal{R}_{th}=90\%$) of the original synthetic samples introduces the noise into the global model's training data, thus affecting the accuracy of the model. Conversely, if the retention ratio is too low ($\mathcal{R}_{th}=40\%$), it leads to insufficient training data and reduced data diversity, affecting the generalization of the final global model. 

Additionally, choosing $\mathcal{R}_{th}$ within the range of 60\%-90\% has little impact on FedMHO-MD and FedMHO-SD, highlighting the robustness of our methods. Specifically, for FedMHO-MD, when the client data distribution satisfies $Dir(0.5)$, $Dir(0.3)$, and $Dir(0.1)$ partitions, the accuracy of the global model varies by 0.41\%, 0.21\%, and 0.08\%, respectively. Similarly, for FedMHO-SD, under the same distributions, the accuracy varies by 0.38\%, 0.24\%, and 0.33\%, respectively.

\begin{table*}[t]
    \normalsize
    \caption{Top-1 test accuracy (\%) of FedAvg, FedDF, DENSE, Co-Boosting, and FEDCVAE when training with homogeneous large local models, and Top-1 test accuracy (\%) of FedMHO-MD and FedMHO-SD when training with heterogeneous local models as in Table~\ref{table_performance_comparison}.}
    \label{table_performance_comparison_homo}
    \centering
    \begin{tabular}{c|c|ccccccc}
    \toprule
        Dataset & Partition & FedAvg & FedDF & DENSE & Co-Boosting & FEDCVAE & FedMHO-MD & FedMHO-SD\\
        \cmidrule(lr){1-9}
        \multirow{3}{*}{MNIST} 
        & $Dir(0.5)$ & 88.44$\pm$1.08 & 91.13$\pm$0.25 & 93.56$\pm$0.74 & 93.61$\pm$0.72 & 94.14$\pm$0.61 & \textbf{95.71$\pm$0.68} & \underline{95.67$\pm$0.47}\\
        
        & $Dir(0.3)$ & 79.96$\pm$1.67 & 86.00$\pm$0.64 & 91.63$\pm$0.44 & 91.95$\pm$0.37 & 93.72$\pm$0.35 & \underline{93.98$\pm$0.69} & \textbf{94.25$\pm$0.18}\\
        
        & $Dir(0.1)$ & 56.67$\pm$2.38 & 78.06$\pm$1.01 & 74.99$\pm$0.29 & 82.34$\pm$0.42 & \textbf{93.01$\pm$0.63} & 91.22$\pm$0.56 & \underline{91.55$\pm$0.27}\\

        \cmidrule(lr){1-9}
        \multirow{3}{*}{Fashion}
        & $Dir(0.5)$ & 67.13$\pm$1.43 & 68.63$\pm$0.78 & 80.59$\pm$1.34 & \textbf{80.82$\pm$0.71} & 76.18$\pm$0.54 & 77.27$\pm$0.48 & \underline{77.73$\pm$0.59}\\
        
        & $Dir(0.3)$ & 54.16$\pm$1.22 & 62.70$\pm$1.31 & 67.18$\pm$1.69 & 69.11$\pm$0.86 & 74.90$\pm$0.72 & \underline{75.69$\pm$0.93} & \textbf{76.45$\pm$0.53}\\
        
        & $Dir(0.1)$ & 38.39$\pm$1.85 & 51.10$\pm$1.98 & 49.26$\pm$0.84 & 55.60$\pm$0.76 & \textbf{74.11$\pm$1.46} & \underline{71.04$\pm$0.50} & 69.65$\pm$0.99\\

        \cmidrule(lr){1-9}
        \multirow{3}{*}{SVHN}
        & $Dir(0.5)$ & 64.14$\pm$0.87 & 75.33$\pm$0.97 & 79.99$\pm$0.53 & 79.56$\pm$0.44 & 85.02$\pm$0.45 & \textbf{85.16$\pm$0.73} & \underline{84.60$\pm$0.71}\\
        
        & $Dir(0.3)$ & 56.62$\pm$1.75 & 73.04$\pm$0.80 & 73.69$\pm$0.77 & 74.08$\pm$0.51 & \underline{83.82$\pm$0.60} & \textbf{84.02$\pm$0.46} & 83.11$\pm$0.06\\
        
        & $Dir(0.1)$ & 45.89$\pm$2.34 & 59.72$\pm$1.25 & 65.16$\pm$1.60 & 69.24$\pm$0.64 & \textbf{83.71$\pm$0.49} & \underline{81.09$\pm$0.56} & 80.53$\pm$0.30\\

        \cmidrule(lr){1-9}
        \multirow{3}{*}{EMNIST}
        & $Dir(0.5)$ & 68.55$\pm$1.46 & 71.07$\pm$0.90 & 80.84$\pm$1.16 & \textbf{81.13$\pm$0.65} & 77.67$\pm$0.78 & \underline{78.45$\pm$0.27} & 77.79$\pm$0.45\\
        
        & $Dir(0.3)$ & 62.03$\pm$0.71 & 66.81$\pm$1.11 & 75.45$\pm$1.93 & 75.96$\pm$0.85 & \textbf{76.48$\pm$0.34} & 74.81$\pm$0.56 & \underline{75.36$\pm$0.51}\\
        
        & $Dir(0.1)$ & 44.16$\pm$0.56 & 49.13$\pm$1.07 & 64.98$\pm$2.68 & 68.39$\pm$1.06 & \textbf{74.55$\pm$0.14} & 73.23$\pm$0.86 & \underline{73.44$\pm$0.78}\\

        \bottomrule
    \end{tabular}
\end{table*}

\subsection{Homogeneous Local Model}
\label{appendix_homogeneous_local_model}
To verify the effectiveness and practicality of our methods, we use the homogeneous large models listed in Table~\ref{table_local_model} to train the baseline methods, with the experimental results presented in Table~\ref{table_performance_comparison_homo}. We observe that replacing small models with large models improves the Top-1 test accuracy of the baselines. Among these methods, FEDCVAE achieves the best performance across various datasets when $Dir(\alpha)$ is set to 0.1. This improvement is due to each client's local data largely contains only a subset of categories. This configuration facilitates the local training of a large generative model that generates high-quality synthetic data.

Despite half of the clients in our methods utilizing small local generative models, our proposed FedMHO-MD and FedMHO-SD achieve comparable or superior performance to the baselines that exclusively employ large local models. For example, when the client data distribution follows $Dir(0.5)$ partition, FedMHO-MD achieves the highest Top-1 test accuracy on the MNIST dataset, while FedMHO-SD achieves the highest accuracy on the SVHN dataset. Both methods achieve the second-highest Top-1 test accuracy on the Fashion and EMNIST datasets, with differences of only 3.09\% and 2.68\% compared to the Co-Boosting method, which is also trained with the large model. Therefore, we conclude that our methods enable resource-constrained clients to participate in FL training and help achieve a high-performing global model. Considering that real-world FL scenarios often involve model heterogeneity, our method demonstrates significant practicality.

\subsection{Extensions for Various Classification Models}
To further demonstrate the scalability of our proposed methods, we evaluate their performance using a more diverse set of classification model prototypes. Within this context, the small models remain constant, and the large models include EfficientNet-V2~\cite{tan2021efficientnetv2}, ResNet-50~\cite{he2016deep}, MobileNet-V2~\cite{sandler2018mobilenetv2}, and VGG-9. Each model prototype is deployed on 2 clients. Each method utilizes 8 larger models and 2 smaller models in local training. The local optimizer for classification models uses Adam with a learning rate of 5$e$-4. To adapt to this scenario, our proposed FedMHOs employ ensemble distillation for global model initialization. FedAvg and FedDF depend on the voting outcomes of each model prototype, while other methods adopt EfficientNet-V2 as the global model and report its accuracy. All other experimental configurations remain consistent. The detailed results on the EMNIST dataset are provided in Table~\ref{table_performance_comparison_2}.

The results indicate that FedMHO-MD and FedMHO-SD consistently outperform the baselines in Top-1 test accuracy. For example, under a $Dir(0.5)$ partition of local data, FedMHO, FedMHO-MD, and FedMHO-SD outperform FEDCVAE, the best baseline, by 5.04\%, 6.77\%, and 7.26\%, respectively. Similarly, for a $Dir(0.3)$ partition, FedMHO, FedMHO-MD, and FedMHO-SD surpass Co-Boosting, the best-performing baseline, by 1.05\%, 4.62\%, and 4.30\%, respectively. Furthermore, under the $Dir(0.1)$ partition, FedMHO-MD and FedMHO-SD outperform FEDCVAE by 2.19\% and 1.16\%, respectively.

\begin{table}[t]
    \normalsize
    \caption{Top-1 test accuracy (\%) comparison on EMNIST dataset under various classification models.}
    \label{table_performance_comparison_2}
    \centering
    \begin{tabular}{c|ccc}
    \toprule
        Method & $Dir(0.5)$ & $Dir(0.3)$ & $Dir(0.1)$ \\

        \cmidrule(lr){1-4}
        FedAvg & 62.38$\pm$2.03 & 50.71$\pm$2.43 & 
        33.07$\pm$1.57  \\
        
        FedDF & 
        68.49$\pm$0.84 & 
        61.65$\pm$0.91 &
        48.98$\pm$1.68 \\

        DENSE &
        72.57$\pm$0.36 &
        70.53$\pm$0.72 & 
        55.76$\pm$1.35 \\

        Co-Boosting &
        72.81$\pm$0.48 &
        71.14$\pm$0.53 & 
        58.23$\pm$0.69 \\

        FEDCVAE & 
        73.12$\pm$1.12 &
        68.38$\pm$1.55 &
        66.03$\pm$2.34 \\
        
        FedMHO &
        78.16$\pm$0.50 & 
        72.19$\pm$1.47 & 
        62.04$\pm$1.73 \\

        \cmidrule(lr){1-4}
        FedMHO-MD & 
        \underline{79.89$\pm$0.55} &
        \textbf{75.76$\pm$1.17} &
        \textbf{68.22$\pm$1.35} \\
        
        FedMHO-SD & \textbf{80.38$\pm$0.24} &
        \underline{75.44$\pm$0.98} &
        \underline{67.19$\pm$1.08} \\
        
        \bottomrule
    \end{tabular}
\end{table}

\begin{figure*}[t]
    \centering
    \subfigure[MNIST]{
\includegraphics[width=0.31\textwidth]{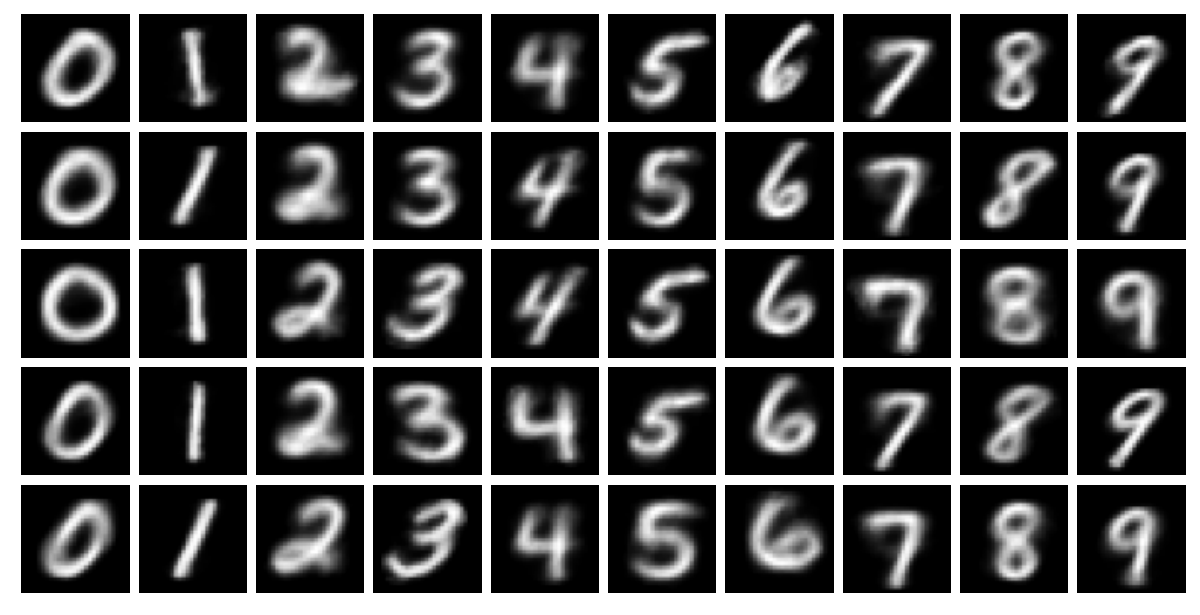}
    }
    \subfigure[Fashion]{
\includegraphics[width=0.31\textwidth]{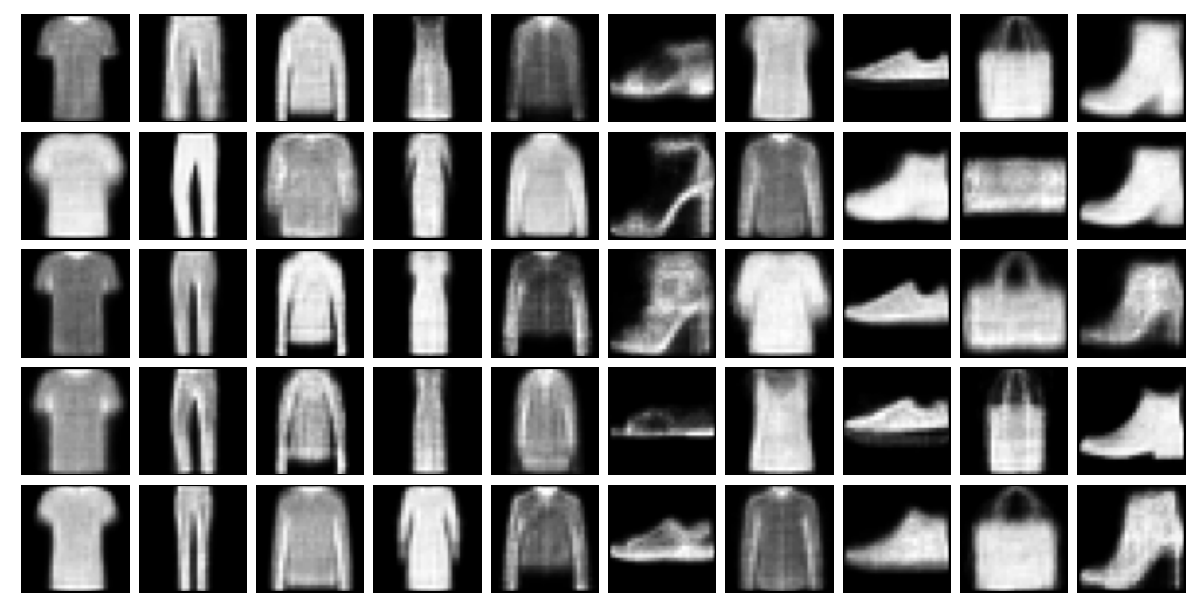}
    }
    \subfigure[SVHN]{
\includegraphics[width=0.31\textwidth]{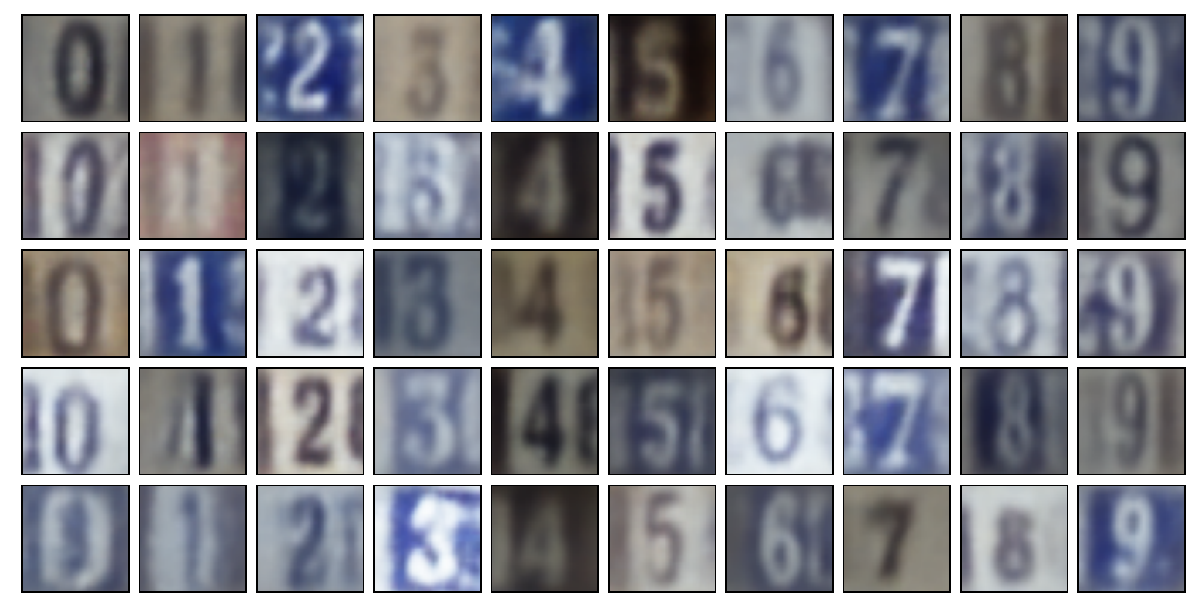}
    }
    
    \subfigure[EMNIST]{
    \includegraphics[width=0.98\textwidth]{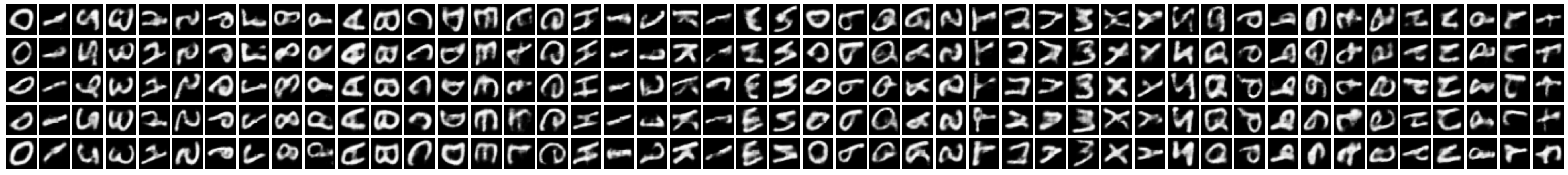}
    }
    
    \caption{Visualization of synthetic samples after unsupervised data optimization. 5 synthetic samples are present for each category.
}
    \label{figure_visualization_synthetic_samples}
\end{figure*}

\subsection{Visualization of Synthetic Samples}
\label{visualization_of_synthetic_data}

We present the visualization of synthetic samples produced by our methods in Figure~\ref{figure_visualization_synthetic_samples}. The use of a lightweight CVAE causes the contours of the generated images to appear somewhat blurred compared to real data. However, the category information remains accurate, with noticeable diversity among samples within the same category. This demonstrates that, although the generated images are less detailed than real data, they reliably represent category information.

\subsection{The Trade-off Between Synthetic Sample Similarity and Privacy}

Balancing the fidelity of synthetic samples with the risk of local privacy leakage is crucial in FL.
We analyze the similarity between synthetic and real training data to illustrate the effectiveness of our methods in enhancing global model accuracy while preserving privacy. To quantify similarity, we use the widely accepted Fréchet Inception Distance (FID) metric~\cite{heusel2017gans}, where lower FID values indicate closer resemblance between synthetic and real data distributions.

\begin{table}[t]
    \normalsize
    \caption{The relationship between the FID of synthetic samples and the global model Top-1 accuracy (\%).  The experiment is conducted on the EMNIST dataset, and the local data partition follows a $Dir(0.5)$ distribution The symbol $E_k$ represents the number of CVAE's local training epochs.}
    \label{table_trade_off}
    \centering

    \begin{tabular}{c|ccc}
    \toprule
         & FID & FedMHO-MD & FedMHO-SD \\

        \cmidrule(lr){1-4}
        Gaussian Noise & 360.53 & N/A & 
        N/A  \\
        
        $E_k=10$ & 145.71 & 67.55\% &
        67.48\% \\

        $E_k=20$ & 127.91 & 74.93\% &
        74.61\% \\

        $E_k=30$ & 110.65 & 77.14\% &
        76.40\% \\

        $E_k=40$ & 102.51 & 77.75\% &
        77.18\% \\

        $E_k=50$ & 100.01 & 78.45\% &
        77.79\% \\

        Real Data & $2e^{-6}$ & N/A & 
        N/A  \\
        
        \bottomrule
    \end{tabular}
\end{table}

Table~\ref{table_trade_off} presents the global model's performance across varying local training epochs $E_k$, alongside the corresponding FID values between synthetic and real samples. Among them, the ``Gaussian Noise" row represents the FID value between Gaussian noise and real data, serving as the upper bound of the FID indicator in this experimental setup. Conversely, the ``Real Data" shows the FID value obtained by splitting the EMNIST dataset into two parts, representing the lower bound in this scenario. As observed in Table~\ref{table_trade_off}, FID values range between 100 and 150 when local training epochs vary from 10 to 50. This result indicates that the generated samples effectively contribute to the global model training while safeguarding privacy. The bias in synthetic samples prevents local data leakage while maintaining competitive global model performance. Notably, when $E_k \geq 30$ (corresponding to $FID \leq 110.65$), synthetic data diverges sufficiently from real data, ensuring privacy protection without compromising global model performance.

\section{Conclusion}

In this paper, we propose a novel one-shot FL method called FedMHO, which aims to enhance the global model's performance when clients deploy local models of varying sizes. FedMHO involves a data generation stage and a knowledge fusion stage to aggregate deep classification models and lightweight generative models. Moreover, we provide an unsupervised data solution to improve the quality of synthetic samples and propose two strategies, FedMHO-MD and FedMHO-SD, to mitigate knowledge forgetting. Extensive experiments conducted across diverse datasets and data partitions validate the efficacy of our method. Overall, FedMHO represents a highly practical framework for performing data-free one-shot FL for computing resource-constrained clients. A promising future direction is to explore potential privacy attacks in one-shot FL.


%

\ifCLASSOPTIONcaptionsoff
  \newpage
\fi



\bibliographystyle{IEEEtran}
\bibliography{reference}

\end{document}